\documentclass[journal]{IEEEtran}
\usepackage{xr}
\usepackage{amsmath,amsfonts}
\usepackage{array}
\usepackage[caption=false,font=normalsize,labelfont=sf,textfont=sf]{subfig}
\usepackage{graphicx}
\graphicspath{{./figures/}}
\usepackage{balance}
\usepackage{hyperref}
\usepackage{xspace}
\usepackage{placeins}
\usepackage{mathtools}

\usepackage[capitalize]{cleveref}
\Crefname{figure}{Fig.}{Figs.}
\crefname{section}{Section}{Sections}
\Crefname{table}{Table}{Tables}
\Crefname{equation}{}{Equations}

\newcommand{\taucamera}{\href{https://www.flir.com/products/tau-2/}{\texttt{Tau2}}}
\newcommand{\blackbody}{\href{https://www.ci-systems.com/sr-800n-superior-accuracy-blackbody}{\texttt{SR-800N}}}
\newcommand{\scientificCamera}{\href{https://www.flir.com/products/a655sc/}{\texttt{A655sc}}}

\newcommand{\tamb}{t_\mathit{amb}}
\newcommand{\nChannels}{\mu}
\newcommand{\scaleFactor}{\xi}
\newcommand{\nFrames}{N}
\newcommand{\tobj}{t_\mathit{obj}}
\newcommand{\mat}[1]{\underline{\underline{#1}}}
\newcommand{\ir}{\textit{IR}\xspace}
\newcommand{\kernels}{\mathcal{K}\xspace}
\newcommand{\temperatureEstimationPerPixel}{\Hat{X}^p\xspace}
\newcommand{\temperatureEstimation}{\Hat{X}\xspace}

\begin{document}
\title{Simultaneous temperature estimation and nonuniformity correction from multiple frames}
\author{\href{mailto:navotoz@mail.tau.ac.il}{Navot Oz},
        \href{mailto:omriberman@mail.tau.ac.il}{Omri Berman},
        \href{mailto:sochen@math.tau.ac.il}{Nir Sochen},
        \href{mailto:mend@eng.tau.ac.il}{David Mendelovich},
         and \href{mailto:iftach@volcani.agri.gov.il}{Iftach Klapp}
         
\thanks{\href{mailto:navotoz@mail.tau.ac.il}{Navot Oz}, \href{mailto:omriberman@mail.tau.ac.il}{Omri Berman} and \href{mailto:iftach@volcani.agri.gov.il}{Iftach Klapp} are with the Department of Sensing, Information and Mechanization Engineering, Agricultural Research Organization, Volcani Institute, P.O. Box 15159, Rishon LeZion 7505101, Israel.}%
\thanks{\href{mailto:navotoz@mail.tau.ac.il}{Navot Oz}, \href{mailto:omriberman@mail.tau.ac.il}{Omri Berman}
        and \href{mailto:mend@eng.tau.ac.il}{David Mendelovich} are with the School of Electrical Engineering, Tel Aviv University, Tel Aviv 69978, Israel.}%
\thanks{\href{mailto:sochen@math.tau.ac.il}{Nir Sochen} is with the Department of Mathematics, Tel Aviv University, Tel Aviv 69978, Israel.}%
\thanks{This work was supported by the Israeli Ministry of Agriculture’s Kandel Program (grant number 20-12-0018).}}


\maketitle
\begin{abstract}
    \ir cameras are widely used for temperature measurements in various applications, including agriculture, medicine, and security.
    Low-cost \ir cameras have the immense potential to replace expensive radiometric cameras in these applications; however, low-cost microbolometer-based \ir cameras are prone to spatially variant nonuniformity and to drift in temperature measurements, which limit their usability in practical scenarios.
    
    To address these limitations, we propose a novel approach for simultaneous temperature estimation and nonuniformity correction (NUC) from multiple frames captured by low-cost microbolometer-based \ir cameras. We leverage the camera's physical image-acquisition model and incorporate it into a deep-learning architecture termed kernel prediction network (KPN), which enables us to combine multiple frames despite imperfect registration between them. We also propose a novel offset block that incorporates the ambient temperature into the model and enables us to estimate the offset of the camera, which is a key factor in temperature estimation.
    
    Our findings demonstrate that the number of frames has a significant impact on the accuracy of the temperature estimation and NUC\@. 
    Moreover, introduction of the offset block results in significantly improved performance compared to vanilla KPN\@. 
    The method was tested on real data collected by a low-cost \ir camera mounted on an unmanned aerial vehicle, showing only a small average error of $0.27-0.54^\circ C$ relative to costly scientific-grade radiometric cameras.
    
    Our method provides an accurate and efficient solution for simultaneous temperature estimation and NUC, which has important implications for a wide range of practical applications.
\end{abstract}

\FloatBarrier
\begin{figure}[t]
    \centering
    \includegraphics[width=1\linewidth]{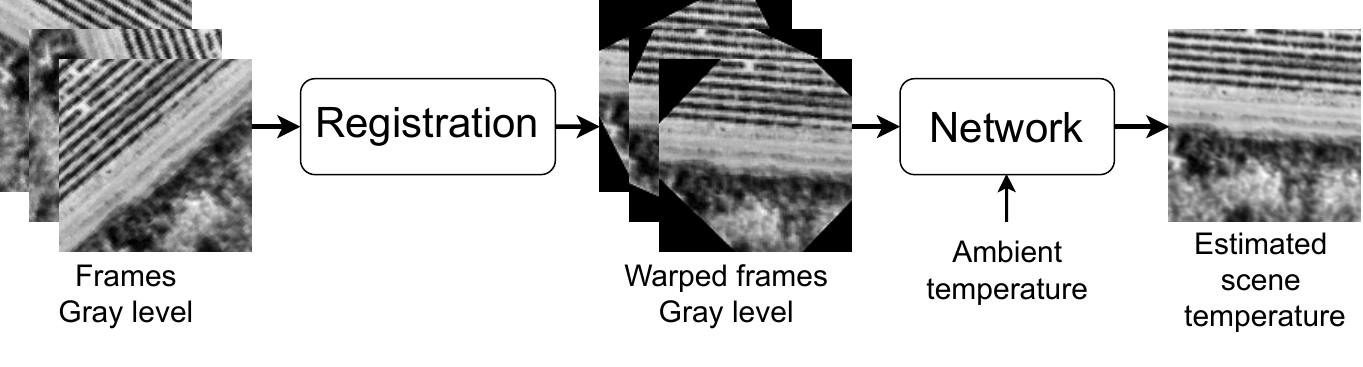}
    \caption{Estimating the scene temperature from a burst of gray-level frames.}
    \label{fig:process}
\end{figure}

\section{Introduction}\label{sec:intro}
    \IEEEPARstart{T}{emperature} is an important indicator of an object's state.
For example, the temperature of a plant is important in deducing information on its well-being~\cite{ir_importance_1, ir_importance_2}. 
Long-wave \ir (LWIR) imaging, commonly termed \ir imaging, measures the thermal radiation emitted from an object. 
To avoid noise and improve accuracy, radiometric \ir cameras employ either a cooling mechanism or a sophisticated shuttering apparatus. 
Both are expensive and energy consuming, which result in a highly expensive camera.
Although \ir imaging is a well-established technique, the high cost of \ir cameras prohibits its widespread use.
There exists an alternative approach to radiometric thermal imaging involving the use of low-cost uncooled microbolometer arrays, which can facilitate the creation of inexpensive \ir cameras with low energy requirements, but with a significant loss in accuracy.
Unlike photon-counting detector arrays, microbolometer arrays gauge alterations in electrical resistance resulting from the radiation emitted from an object~\cite{bolometer}. Each microbolometer in the array is heated by the thermal radiation to a temperature that is dependent on the scene, resulting in each microbolometer having a marginally different temperature based on the observed scene and the incident angle of the radiation. The incident radiation causes a miniscule change in the resistance of the microbolometer. The temperature of the scene is reflected by the variation in resistance of each microbolometer. The infinitesimal changes in resistance detected by each microbolometer are used to create an image that corresponds to the temperature of the observed scene.

Although microbolometer arrays are a useful tool for thermal imaging, they have significant limitations. Space-variant nonuniformity and noise from various sources affect the accuracy of these arrays. The nonuniformity drifts due to the change in ambient temperature, which causes unpredictable errors in the sensor readings.
The lack of a cold shield in the uncooled camera is a prominent cause of nonuniformity~\cite{IrFundamentals}. This self-radiation effect is attributed to the camera's housing and lens, which emit thermal radiation onto the sensor. This self-radiation varies according to the ambient temperature of the camera.

\begin{figure}
    \centering
    \subfloat{\includegraphics[width=0.49\linewidth]{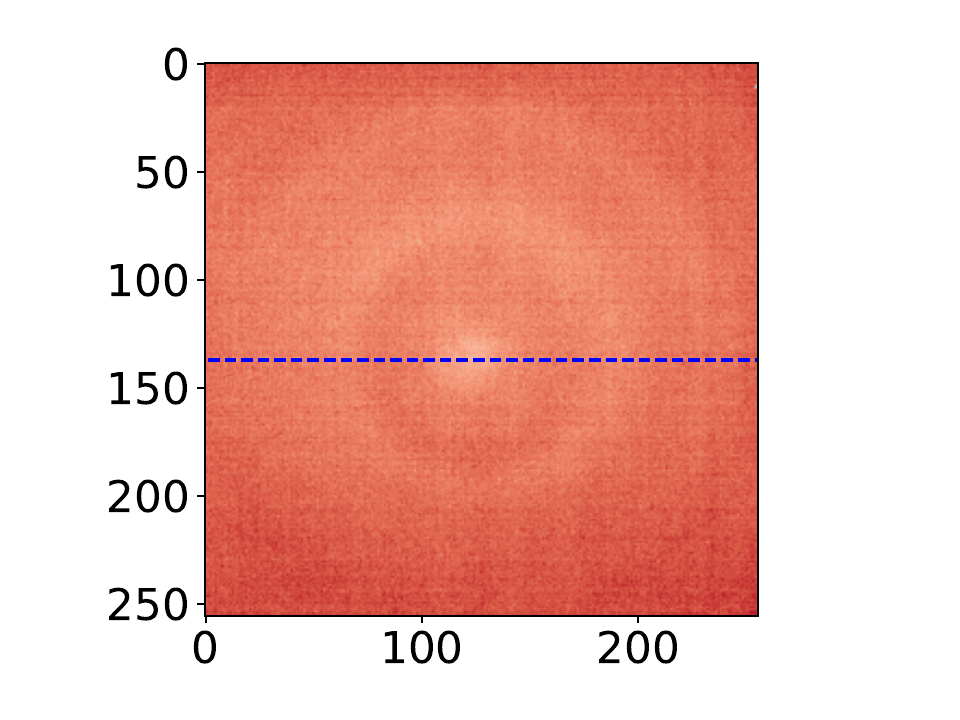}}%
    \hfill%
    \subfloat{\includegraphics[width=0.49\linewidth]{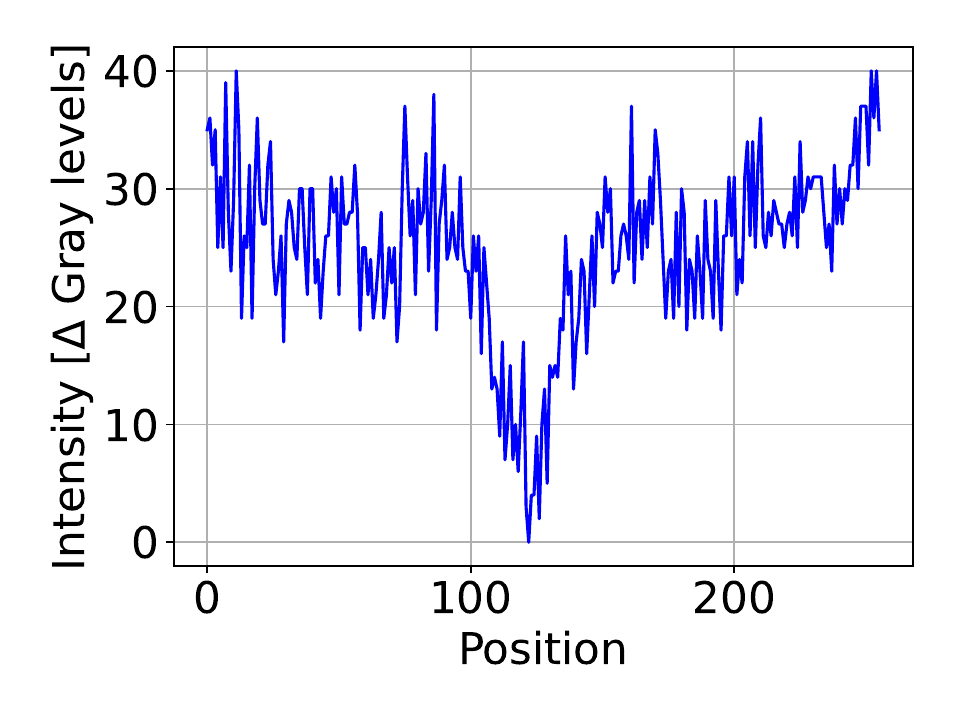}}%
    \caption{Example of the nonuniformity in low-cost \ir cameras. On the left is an image of a $30^\circ C$ blackbody with ambient temperature of $44.7^\circ C$, and on the right are the intensities along the blue dashed line.}
    \label{fig:nonuniformity}
\end{figure}

Fixed-pattern noise (FPN) is another factor that contributes to nonuniformity in microbolometer arrays. The readout circuitry of these arrays is typically line-based, like charge-coupled devices. Even minor differences between line-readers on the same array can result in significant variation between lines in the resulting image \cite{Riou2004}.
Noises in the camera increase the noise equivalent differential temperature (NEDT), which refers to the minimum detectable change in scene temperature~\cite{Riou2004}. The NEDT is a measure of the sensitivity of the camera; the higher the NEDT, the less sensitive the camera is to changes in temperature.
An image of a uniform heat source (blackbody) is shown in \cref{fig:nonuniformity}. The spatially-variant nonuniformity is demonstrated by the radial patterns in the gray levels of the left subfigure. The subfigure on the right plots the gray levels along the blue dashed line, showing the impact of nonuniformity and noise on the gray levels.

A widely used application of \ir imaging is remote sensing - the process of acquiring information about an object without making physical contact with it. The information is acquired by measuring the radiation that is reflected by, or emitted from the object. The information is then used to deduce its physical properties. Remote sensing is used in a variety of fields (e.g., agriculture, geology, and meteorology).

One common use for \ir cameras is to mount them on drones. 
This setup results in high overlap between frames (\cref{sec:results:realdata}). 
The redundant information can be used to simultaneously improve the accuracy of the temperature estimation and correct nonuniformity in the frames.
\cref{fig:approach:redundentNonUniformity} illustrates how redundant information between frames can be beneficial. The object is affected differently by the nonuniformity in each frame, which means that the true underlying temperature of the object can be extracted.

The aims of this study are to: exploit the redundancy in the data and the physical model of the camera to develop a method of estimating scene temperatures using a low-cost microbolometer-based \ir camera, and correct for nonuniformity in the frames.
\section{Related work}\label{sec:prior}
    \noindent The estimation of temperature can be broadly divided into two parts: transforming the output of the camera to temperatures, and correcting nonuniformity in the sensor.
Determining the transformation from camera output to temperature is called \emph{thermal calibration}. Correcting the nonuniformity in the sensor is called \emph{nonuniformity correction}~(NUC).

\subsection{Thermal calibration}\label{sec:prior:thermalDrift}
    \noindent The raw output of the \ir camera is dependent on the object temperature, and the output values themselves are given in gray levels.
For example, the dynamic range of the gray level in the low-cost \ir \taucamera\ is $14_{bits}$. 
The classical approach is to calibrate the camera for different ambient temperatures~\cite{Schulz1995}. 

A large dataset of object-ambient temperature pairs must be collected for calibration.
The gain and offset are calculated from the per-pixel data to determine the spatially variant nonuniformity. Thus, the calibration process usually requires considerable time and resources.

Schulz and Caldwell~\cite{Schulz1995} used a single-point correction, i.e., a single ambient temperature was used, a constant gain was assumed, and only the offset was found. 
Riou et al.~\cite{Riou2004} suggested a two-point correction that requires two ambient temperatures, but solved for both gain and offset; this correction is widely used across industrial \ir cameras today.
Both methods use a linear regression to extract the gain and offset coefficients.
Nugent et al.~\cite{Nugent2013} modeled the gain and offset as a polynomial in the temperature of the object and used least-squares to extract the coefficients.
Contemporary work adds prior knowledge to the calibration process. 
Liang et al.~\cite{Liang2017} found the gain and offset for a given temperature and interpolated the results for other ambient temperatures. Chang and Li~\cite{Chang2019} incorporated the integration time of each frame as prior knowledge to the calibration.

The calibration data must be collected for each camera separately, because each camera is slightly different due to the manufacturing process. This requires scientific-grade equipment, making the calibration process infeasible for most users.

\subsection{Nonuniformity correction}\label{sec:prior:nuc}
    \begin{figure}
    \newcommand{\sizeRedundantRect}{0.35}
    \centering
    \begin{minipage}[c]{0.4\linewidth}
    \begin{flushleft}
        \subfloat{\includegraphics[width=\linewidth]{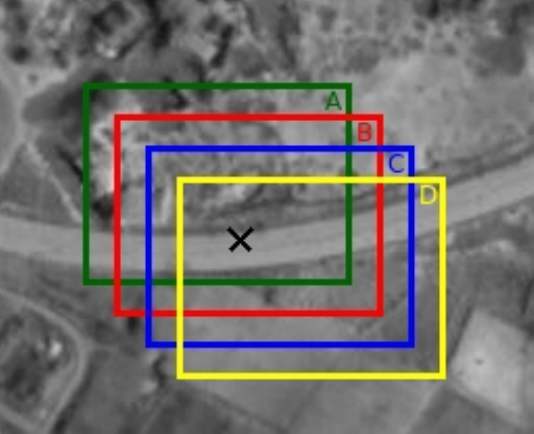}}%
    \end{flushleft}
    \end{minipage}
    \hfill
    \begin{minipage}[c]{0.55\linewidth}
    \begin{flushright}
        \begin{minipage}[c]{\linewidth}
            \subfloat{\includegraphics[width=\sizeRedundantRect\linewidth]{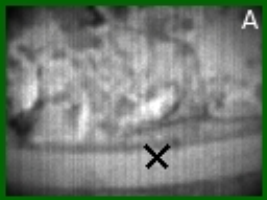}}
            \hfill
            \subfloat{\includegraphics[width=\sizeRedundantRect\linewidth]{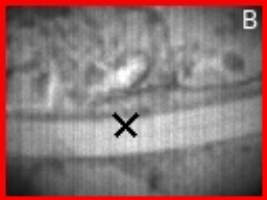}}
        \end{minipage}
    \end{flushright}
    \centering
    \begin{minipage}[c]{\linewidth}
            \subfloat{\includegraphics[width=\sizeRedundantRect\linewidth]{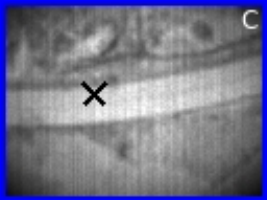}}
            \hfill
            \subfloat{\includegraphics[width=\sizeRedundantRect\linewidth]{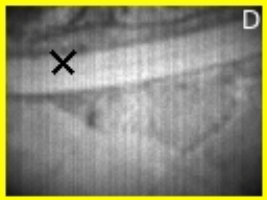}}
        \end{minipage}
    \end{minipage}
    \caption{Simulation of consecutive frames taken during a drone flight from an \ir camera. 
    The frames sampled on the left image are marked by colored rectangles. 
    The effects of the spatially variant nonuniformity are seen in the frames on the right.
    The cross on the road appears in different locations in the different frames and is affected differently by the spatially variant nonuniformity.}
    \label{fig:approach:redundentNonUniformity}
\end{figure}
\noindent As stated in \cref{sec:intro}, the frames of the \ir camera suffer from spatially variant nonuniformity. The nonuniformity can be corrected for a single frame, or by combining information from multiple frames (known as scene-based).

\subsubsection{Single Frame}\label{sec:prior:nuc:singleFrame}
A given image contains information that can be exploited for different tasks, such as low frequencies~\cite{Oz2020}, recurring patches in the image~\cite{Shocher2017} or the statistical distribution of patches in the image~\cite{Shaham2019}.
Some works used a single image to correct the nonuniformity.

Scribner et al.~\cite{Scribner91} used a neural network (NN) to find the offset and gain by alternating optimization and gradient descent.
Tendero and Gilles~\cite{Tendero12} used histogram equalization across the columns in a frame, and then applied a discrete cosine transform to denoise the frame.
Cao and Tisse~\cite{Cao2014} relied on spatial dependence between adjunct pixels to estimate both the ambient temperature and the correction.
Zhao et al.~\cite{Zhao13} solved an optimization problem, with a constraint on the directional gradients of each frame.

Recent work has applied deep learning (DL) methods for single-image NUC\@.
Jian et al.~\cite{Jian2018} learned the nonuniformity pattern from the filtered high frequencies of the frames.
He et al.~\cite{He2018} trained a convolutional neural network (CNN) that outputs a corrected image end to end (E2E).
Chang et al.~\cite{ChangDeepLearning2019} constructed a multiscale network to reconstruct a corrected frame.
Saragadam et al.~\cite{Saragadam2021} solved an optimization problem with a NN as the prior, and a physical model as the constraint.
Oz et al.~\cite{Oz2022} modeled the nonuniformity and trained a network based on the physics of the acquisition model.

Single-image methods require only a single frame so they are easier to apply, but their performance is degraded compared to scene-based methods.

\subsubsection{Scene-Based}
Scene-based studies rely on the assumption that the change in ambient temperature is slower than the frame rate, and therefore the gain and offset are constant between consecutive frames.

Harris and Chiang~\cite{Harris99} calculated shift and normalization terms per pixel and updated these terms recursively when new frames arrived.
Hardie et al.~\cite{Hardie00} registered the frames and then averaged the results per pixel.
Vera and Torres~\cite{Vera05} improved the NN suggested by Scribner et al.~\cite{Scribner91} with an adaptive learning rate and a different loss function that accounts for multiframe information.
Averbuch et al.~\cite{Averbuch2007} reformulated the NUC problem to a Kalman filter.
Zuo et al.~\cite{Zuo11} estimated per-pixel \textit{irradiance} between two frames.
Papini et al.~\cite{Papini2018} approximated the gain and offset from multiple pairs of blurred and sharp images.
The common characteristic of these studies is that an update step must be performed when new frames arrive, before the correction step. The combined update and correction steps are computationally intensive and pose a constraint on the run time of the system.

A NN-based method to simultaneously estimate the scene temperature and correct the nonuniformity using multiframe information has not yet been achieved.

The present study builds on the image-acquisition model, which describes the relationship between the observed scene and the output of the camera (\cref{sec:background:imageAcquasition}).
By leveraging redundant information across multiple frames and ambient temperature data, the study develops a kernel prediction network (KPN) that uses DL techniques to estimate the temperature of each pixel (\cref{sec:methods:net}).

The efficacy of the method is demonstrated through comparisons of real measurements obtained with an uncooled \ir camera and those from a scientific radiometric camera.
These tests illustrate the method's ability to correct for nonuniformity and estimate temperatures accurately across different cameras (\cref{sec:results:realdata}).

Our main contributions consist of: (1) exploiting the redundant information between frames to simultaneously estimate the scene temperature and correct the nonuniformity using a NN; (2) imposing the physical model of the camera as a constraint on the network to enhance the temperature-estimation accuracy; (3) incorporating the ambient temperature data as an additional input to the network to further improve the accuracy of the temperature estimation; and (4) demonstrating the advantages of using multiple frames over single-frame methods through extensive experiments on synthetic and real data.    
\section{Background}\label{sec:background}
    We develop the physical image-acquisition model of the \ir camera in \cref{sec:background:imageAcquasition}, and then expand it to multiple frames in \cref{sec:background:multiframes}.

\subsection{Image acquisition}\label{sec:background:imageAcquasition}
    \noindent A \emph{blackbody} is an ideal Lambertian surface that emits the maximal radiation at any given wavelength.
The spectral density of radiation emitted from a blackbody is described by Planck's law~\cite{IrFundamentals}:
\begin{equation}\label{eq:acquisition:plancksLaw}
    M_\lambda(T) = \frac{2\pi h c^2}{\lambda^5}\frac{1}{\exp(\frac{hc}{\lambda k T})-1}\quad[W\cdot sr^{-1}\cdot m^{-3}]
\end{equation} where $T$ is the temperature of the blackbody in Kelvin, $\lambda$ is the radiation wavelength, $h$ is Planck's constant, $k$ is the Boltzmann constant and $c$ is the speed of light.

The power emitted over the entire bandwidth is found using the Stefan-Boltzmann law~\cite{IrFundamentals}:
\begin{equation}\label{eq:acquisition:boltzmann:ideal}
    M(T)=\int_0^\infty{M_\lambda(T)d\lambda}=\sigma\cdot T^4\quad[W\cdot sr^{-1}\cdot m^{-2}]
\end{equation} where $\sigma$ is the Stefan-Boltzmann constant.

\cref{eq:acquisition:plancksLaw,eq:acquisition:boltzmann:ideal} hold for an ideal blackbody. Real objects can never emit the maximal radiation for a given wavelength due to physical constraints (e.g., material, viewing angle). The ratio between the ideal emission and the practical emission of an object is called \emph{emissivity}.
Thus, the Stefan-Boltzmann law for radiance power of practical objects is:
\begin{equation}\label{eq:acquisition:boltzmann:practical}
    M(T)=\sigma\cdot\epsilon\cdot T^4\quad[W\cdot sr^{-1}\cdot m^{-2}]
\end{equation} where $\epsilon$ is the emissivity.

The incident power by an object on a microbolometer is estimated by integrating over the physical dimensions of the system in \cref{eq:acquisition:boltzmann:practical}. The incident power on the microbolometer can be written as~\cite{IrFundamentals}:
\begin{equation}\label{eq:acquisition:incidentpower}
    \phi(T)=\gamma\cdot\sigma\cdot\epsilon\cdot T^4\quad[W]
\end{equation} where $\gamma$ is a coefficient that accounts for the dimensions of the object and the camera's field of view.

The intensities of the pixels in radiometric \ir cameras (i.e., gray levels) are linearly proportional to the incident power on the microbolometer.
To model the intensities, we consider a small environment near a reference temperature $T_0$ and expand the Stefan-Boltzmann law in \cref{eq:acquisition:incidentpower} a by Taylor series. In Kelvin, the temperature of the object can be considered a small perturbation around a reference temperature, because the reference temperature is usually hundreds of Kelvins, whereas $\Delta T$ is usually tens of Kelvins. The Taylor expansion of \cref{eq:acquisition:incidentpower} is:
\begin{equation}\label{eq:acquisition:affine}
    \begin{split}
        I(\tobj)&=\gamma\epsilon\sigma T^4=\gamma\epsilon\sigma(\Delta T + T_0)^4\\
        &\approx 4\gamma\epsilon\sigma T_0^3\Delta T + \gamma\epsilon\sigma T_0^4\\
        &\approx g\cdot\tobj + d
    \end{split}
\end{equation}
where $I(\tobj)$ is the gray-level output of the \ir camera, and $g=4\gamma\epsilon\sigma T_0^3, d=\gamma\epsilon\sigma T_0^4$ are the gain and offset coefficients, respectively.
Using the relationship between Kelvin and Celsius, we denote $\tobj\equiv T-273.15$ in $^\circ C$.

Equation~\ref{eq:acquisition:affine} shows that the radiation is linear for scene temperature in the small environment near $T_0$, with the terms $g$ and $d$ dependent on the object temperature relative to $T_0$.

The incident power $\phi(\tobj)$ in \cref{eq:acquisition:incidentpower} changes the temperature of the microbolometer by a small fraction. The change in temperature also changes the electrical resistance of the microbolometer~\cite{uncooled_thermal_imaging}.
By applying a constant electrical current on the microbolometer and using an Ohm-like law, a map between the incident power and the voltage of the microbolometer can be derived~\cite{IrFundamentals}.
In a low-cost uncooled \ir camera, the resistance of the microbolometer changes with the ambient temperature.

To account for this, the gain and offset of the \ir camera are modeled as a function of the ambient temperature~\cite{Nugent2013}:
\begin{equation}\label{eq:acquisition:frame}
    I(\tobj,\tamb)  = g(\tamb) \cdot \tobj + d(\tamb)
\end{equation}

For a given ambient temperature, the gray levels of pixel $[u,v]$ can be written as:
\begin{equation}\label{eq:acquisition:affine:pixel}
    I(\tobj)[u,v]=g[u,v]\cdot\tobj[u,v]+d[u,v]
\end{equation}
The gain and offset are two-dimensional (2D), and together they model the space-variant nonuniformity.

The SNR of uncooled \ir cameras is often low due to noises, the most dominant being $\frac{1}{f}$ and electronic (Johnson) noise~\cite[Chapter 5]{uncooled_thermal_imaging}. The $\frac{1}{f}$ noise is more dominant because the camera operates at a low frequency. $\frac{1}{f}$ noise can be modeled as Gaussian~\cite{ir_1f_gaussian} with zero mean.
\subsection{Multiframes}\label{sec:background:multiframes}
    \noindent Consecutive frames over a brief period of time have overlap between them [an example of a real unmanned aerial vehicle (UAV) pattern with overlapping frames can be seen in \cref{sec:results:realdata}]. These consecutive frames are called a burst. The overlap between frames implies that the same object appears in multiple frames. 
As seen in \cref{eq:acquisition:affine}, the gain and offset are dependent on the pixel location on the sensor, thus different views of the same object can be exploited as redundant information. 
The redundant information between frames is demonstrated in \cref{fig:approach:redundentNonUniformity}.
To exploit it, first an object must have the same coordinates across all frames. To achieve coordinate alignment, registration is performed.

Image registration is the process of aligning two or more images of the same scene taken from different viewpoints or at different times~\cite{computationalGeometry}. Transforming a source frame toward the coordinate system of another destination frame is called a projective transformation, or a homography~\cite[Ch.~0]{HartleyRichard2004MVGi}.
Homography transformation preserves co-linearity between the frames.
Moreover, a homography is invertible and linear by definition~\cite[Def.~2.9]{HartleyRichard2004MVGi}. In layman's terms, the homography preserves the shapes and relations between objects. 

After applying the homography on the source frame, an object should have the same coordinates in both source and destination frames. The transformed source frame is called a warped frame. Expanding to $N$ frames, there exists a set of projective transformations $m_1,\ldots,m_N$ toward a common plane such that the overlap between the frames is maximal~\cite[Ch.~4]{HartleyRichard2004MVGi}. Objects that appear in the overlapping area will have the same coordinates in every warped frame. For our practical use, we choose a pivot frame for each burst of frames and annotate the pivot frame as $\mathcal{I}$.

An underlying assumption throughout this work is that the gain and offset in \cref{eq:acquisition:affine:pixel} are constant for a series of frames taken over a short duration of time (a second). This assumption holds because the ambient temperature of the camera changes at a much slower rate (several minutes).

Let $X$ be the fourth power of an accurate 2D temperature map, and $I_1,\ldots,I_N$ be a set of $N$ frames of $X$ captured by the \ir camera. $I_i$ is in gray levels.
$I_i^{x_i,y_i}$ is the value of the pixel in the $[x_i,y_i]$ location of $I_i$. 
$[u,v]$ are the coordinates of the pivot frame $\mathcal{I}$. 
The frames in the burst can be formulated as:
\begin{equation}\label{eq:multiframe:setOfFrames}
    \begin{matrix}
        I_1^{x_1,y_1}=g^{x_1,y_1}(\tamb)\cdot m_1^{-1}\left(X^{u,v}\right)+d^{x_1,y_1}(\tamb)\\
        \vdots\\
        I_N^{x_N,y_N}=g^{x_N,y_N}(\tamb)\cdot m_N^{-1}\left(X^{u,v}\right)+d^{x_N,y_N}(\tamb)
    \end{matrix}
\end{equation} where $m_1,\ldots,m_N$ are the set of homographies that transforms each frame into $\mathcal{I}$. The zero-mean noise $\mathcal{N}$ was omitted for brevity.

Equation~\ref{eq:multiframe:setOfFrames} formulates the acquisition process of a frame as projecting the temperature map $X$ using the inverse of the homography $m_i$, and then sampling the projected $X$ by applying the gain $g$, offset $d$ and noise $\mathcal{N}$. 
Notice that an object will be sampled at different coordinates for each frame, and since the gain and offset are spatially variant, the object will have a different gain and offset for each frame.
The result in \cref{eq:multiframe:setOfFrames} means that an object appearing in pixel $[u,v]$ of the temperature map $X$ will have multiple representation with different values of $g,d$ and $\mathcal{N}$, enabling the use of redundant information between the frames.
Redundant information between frames has been used for many image-restoration tasks, such as super-resolution~\cite{multi_sr_classic, multi_sr_classic2}, denoising~\cite{multi_denoise_classic}, and deblurring~\cite{multi_deblur_classic}. 
Many recent studies in the area have used DL for either alignment~\cite{Detone16} or fusion of frames~\cite{multi_denoise_dl}, or both~\cite{multi_sr_dl1, multi_sr_dl2, multi_sr_dl3}.
\section{Temperature estimation}\label{sec:method}
    \noindent The proposed method simultaneously estimates the scene temperature and corrects nonuniformity from a burst of consecutive frames.
An overview of the method is presented in \cref{fig:process}.
The \ir camera captures overlapping gray-level frames. 
A burst of consecutive gray-level frames is registered toward $\mathcal{I}$ - the pivot frame that has the maximal overlap with all other frames.
The registered gray-level frames are the input to the network, along with the ambient temperature.
The output of the network is a 2D map of the estimated scene temperatures.

\subsection{Network}\label{sec:methods:net}
    \begin{figure}
    \centering
    \includegraphics[width=1\linewidth]{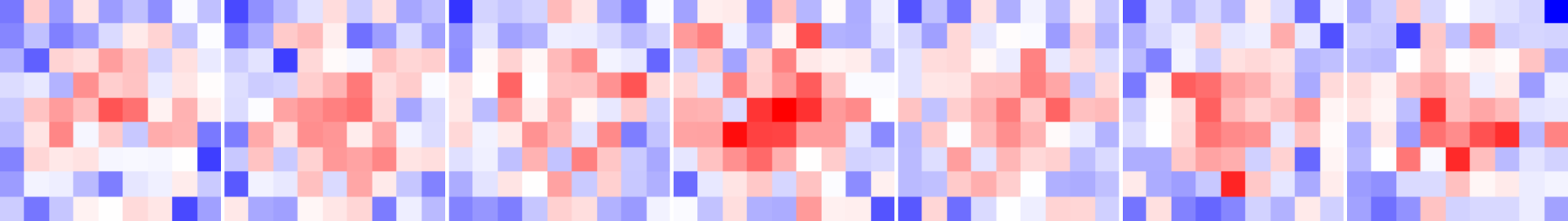}
    \caption{Example of kernels $\kernels$ for a single pixel estimated by the network in \cref{fig:methods:network}, each kernel for a different frame in the burst. The dimensions of the kernels in the figure are $9\times 9$ and they are predicted for the center pixel of $7$ consecutive frames. The middle kernel is from the reference frame. Red has a higher magnitude, and blue has a lower magnitude.}
    \label{fig:methods:kernels}
\end{figure}
\begin{figure}
    \centering
    \includegraphics[width=0.95\linewidth]{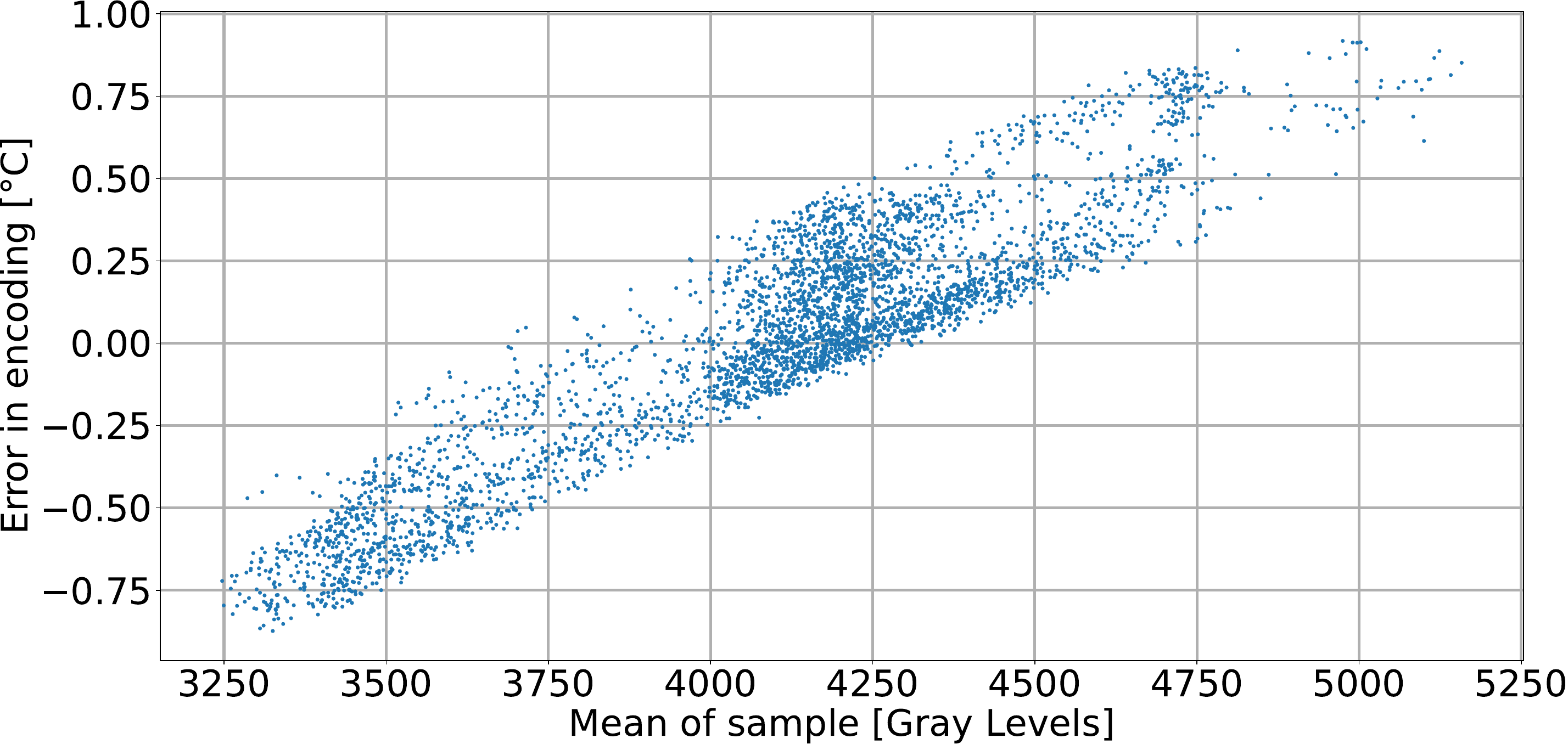}
    \caption{The difference between the true temperatures and the offset block's estimates from \cref{eq:methods:temperature_encoding}, for different average input gray levels.}
    \label{fig:methods:temperature_encoding}
\end{figure}
\noindent In \cref{eq:multiframe:setOfFrames} we show that different views of the same object have usable redundant information.
To exploit the redundancy, these different perspectives require accurate mapping of the frames to pivot frame $\mathcal{I}$.
Na\"ively, a temperature map $\Hat{X}$ can be estimated from \cref{eq:multiframe:setOfFrames} by:
\begin{equation}\label{eq:methods:naiveSolution}
    \begin{split}
        \Hat{X}_{naive}^{u,v} &= \frac{1}{\nFrames}\sum_{i=1}^\nFrames\left[\frac{1}{g^{u-x_i,v-y_i}}\Tilde{I}^{u,v}_i - \frac{d^{u-x_i,v-y_i}}{g^{u-x_i,v-y_i}}\right]\longrightarrow\\
        \Hat{X}_{naive}^{u,v} &= \frac{1}{\nFrames}\sum_{i=1}^\nFrames\left[G^{u-x_i,v-y_i}\Tilde{I}^{u,v}_i + D^{u-x_i,v-y_i}\right]
    \end{split}
\end{equation} where 2D coefficient maps $G$ and $D$.
The na\"ive approach requires exact registration between frames.
The information must be located on the exact same coordinates across all frames.
Inaccurate registration leads to artifacts or ghosting, as well as inexact temperature estimation.
Even with a robust registration framework there is always some degree of misalignment between frames, so the na\"ive approach is unsuitable for practical use.

The method for temperature estimation proposed in this work is robust to misalignment between frames.
The frames are registered toward $\mathcal{I}$ using any off-the-shelf registration method, then fed into a NN that predicts a kernel for each pixel in every frame of the burst.
The kernels are then applied on overlapping patches around each pixel by an inner product between the patch and kernel.
Our method is based on the KPN proposed by De~Brabandere et al.~\cite{DeBrabandere16}.
\cref{fig:methods:kernels} shows kernels predicted by the network. The kernels compensate for misalignment between frames by spatially shifting their center to compensate for the shifts.

The architecture of the temperature estimation network is based on UNET~\cite{unet}, with the kernel prediction block attached to the rear end of the decoder. The encoder and decoder are composed of three $3\times 3$ convolution layers with activations and normalizations, and are described in \cref{table:nn_architecture} in the supplementary material.
The kernel prediction block is composed of three $1\times 1$ convolution layers with activations, and is described in \cref{table:kpnBlock} in the supplementary material.
The entire network architecture is detailed in the supplementary material.

Although the KPN corrects nonuniformity, its temperature estimation is inaccurate.
To improve the latter to match radiometric cameras, we used the ambient temperature as prior information to calibrate the output of the network.
The offset between the gray-level frames and the temperatures was modeled as a polynomial of the mean of the gray-level frames and the ambient temperature:
\begin{equation}\label{eq:methods:temperature_encoding}
    \Tilde{d}\left(\Tilde{I},\tamb\right) = \frac{1}{N}\sum^N_{n=1}\underbrace{\left[\sum_{i,j=0}^\nu\delta_{i,j}\cdot \text{Mean}\left({\Tilde{I}_n}\right)^i \cdot \tamb^j\right]}_{\Tilde{d}_n}
\end{equation} where $\Tilde{d}_n$ is the offset for frame $n$, $\text{Mean}\left({\Tilde{I}_n}\right)$ is the spatial mean of the $n$th gray-level frame, $\tamb$ is the ambient temperature, $\delta_{i,j}$ are the coefficients of the polynomial, and $\nu$ is the degree of the polynomial.

\begin{figure}
    \centering
    \includegraphics[width=1\linewidth]{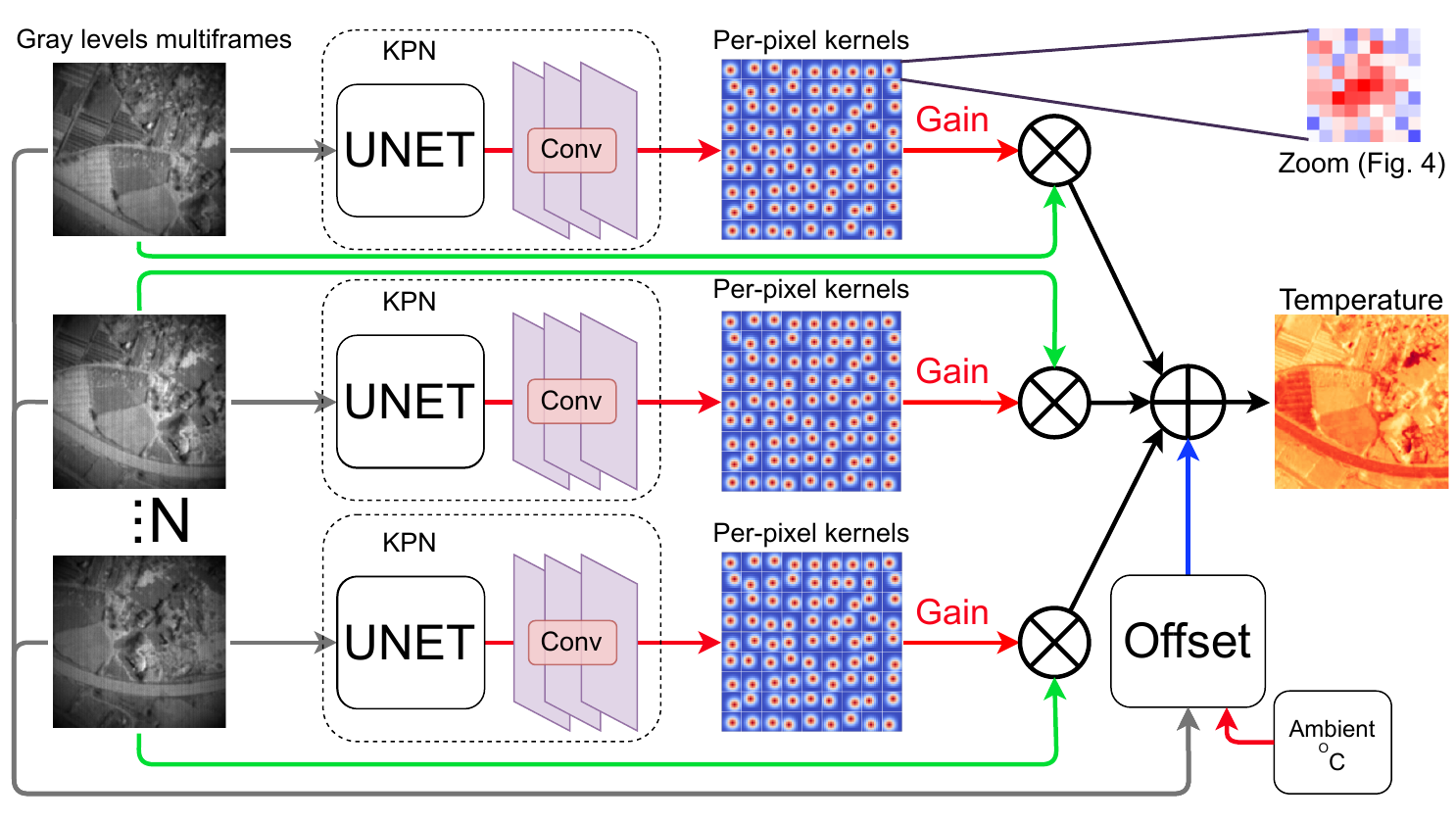}
    \caption{Schematics of the model.
        The gray-level multiframes are fed into the kernel prediction network (KPN), and the KPN outputs the per-pixel kernels $\kernels$ for each frame.
        Each frame is divided into overlapping patches with the same support as the kernels.
        The patches and the kernels are multiplied element-wise and each product is summed, resulting in a 2D gain map for each frame.
        All of the 2D gain maps are summed depth-wise, resulting in a single 2D map.
        The offset, a single scalar value, is added to the single 2D map to get the estimated temperature map.
        A zoom-in example of the kernels $\kernels$ for a single pixel is shown in \cref{fig:methods:kernels}.
        A detailed description of the network architecture and an enlarged figure of the network, \cref{supp_network}, can be found in the supplementary material.}
    \label{fig:methods:network}
\end{figure}

The offset block was jointly trained with the network, allowing the entire network to train end-to-end.
Namely, the coefficients $\delta_{i,j}$ of $\Tilde{d}_n$ were realized by a set of $N\times(\nu+1)$ weights organized in a matrix, such that a single matrix multiplication and summation is required to calculate the offset for all frames.
We found that a polynomial of degree $\nu=4$ offers sufficient improvement in the accuracy of the temperature estimation, and that training the offset block separately from the network does not offer significant improvement.
\cref{fig:methods:temperature_encoding} shows the results of the offset block. The error between the temperature estimation of the offset block and the ground truth (GT) temperature is shown. The error is sub-degree Celsius, and the offset is accurate enough to calibrate the output of the network.

The following equation describes the temperature estimation by applying a KPN to the image-acquisition model.
To combine the information from multiple frames, the gain term in \cref{eq:multiframe:setOfFrames} is generalized as a KPN, and the information from all frames is used.
The kernels applied to each pixel handle the nonuniformity and noise, and the offset term in \cref{eq:methods:temperature_encoding} handles the thermal calibration, resulting in the temperature estimation $\temperatureEstimationPerPixel$:
\begin{equation}\label{eq:methods:KPN}
    \temperatureEstimationPerPixel = \sum_{n=1}^\nFrames\left<\kernels^p_n,S^p\left(\Tilde{I}^p_n\right)\right> + \Tilde{d}\left(\Tilde{I},\tamb\right)
\end{equation} where $\nFrames$ is the number of frames in a burst, $\kernels$ is the kernel of size $K\times K$ produced by the kernel prediction block, and $S(\cdot)$ is a function that crops a $K\times K$ patch around a pixel $p$ in the support of the frames.

The scheme of the model is shown in \cref{fig:methods:network}. The registered burst of frames is fed into the network, which outputs a kernel for each pixel in each frame.
These kernels $\kernels$ serve as the \textit{gain} in \cref{eq:methods:KPN}.
The registered frames are also fed to the offset block along with the ambient temperature, which outputs the \textit{offset} term in \cref{eq:methods:KPN}.
The gain is applied to the frames and the results are summed depth-wise.
The scene temperature estimation $\temperatureEstimation$ is obtained by adding the offset term to the result of the depth-wise summation.
\subsection{Loss functions}\label{sec:methods:losses}
    \noindent The loss is comprised of a fidelity term, a gradient smoothness term, and a structural term.

The structural term $\mathcal{L_{SSIM}}$ maximizes the commonly-used structural similarity metric (SSIM). SSIM loss improves results in image-restoration tasks~\cite{ssimLoss2017}.
The fidelity and gradient terms are similar to Mildenhall et al.~\cite{mildenhall2018kpn}, except that the $L_1$ loss is used instead of $L_2$, because $L_1$ is more robust to outliers~\cite{anwar2020}. The loss function is formulated as:
\begin{equation}\label{eq:methods:loss}
    \begin{split}
        \mathcal{L} = 
        &\left|\left| M(\Hat{X})-M(X) \right|\right|_1+\\
        &\lambda_1\left|\left| M(\nabla\Hat{X})-M(\nabla X) \right|\right|_1+\\
        &\lambda_2\cdot\mathcal{L_{SSIM}}(M(\Hat{X}),M(X))
    \end{split}
\end{equation} where $\Hat{X}$ is the temperature estimated by the network, $X$ is the GT temperature, $M$ is a mask of valid pixels in the registration process, $\lambda_1, \lambda_2$ are hyperparameters to balance to losses, and $\nabla$ is the magnitude of the Sobel operators. The mask is produced by the registration algorithm.
The final values of the hyperparameters were set to $\lambda_1=0.1$ and $\lambda_2=0.01$.
\subsection{Synthetic data}\label{sec:methods:synthData}
    \noindent The network was trained with synthetic data in a supervised manner.
The inputs to the network were created from accurate 2D temperature maps collected using a scientific-grade \ir camera (\scientificCamera).
A degradation model of a low-cost \ir camera (\taucamera) was applied to the temperature maps, transforming them to gray-level frames. As a result, the network was trained on transforming gray-level frames to accurate temperature maps.

The goal of the degradation model was to faithfully transform temperature maps into gray-level maps, allowing the supervised training process of the network. 
The modeling process had three stages:
1) collecting data with the \ir camera in a controlled environment;
2) finding per-pixel coefficients using the image-acquisition model in \cref{sec:background:imageAcquasition};
3) using adjunct pixel dependencies as a constraint on the degradation model.

The degradation model required frames of objects with known temperature taken by the \taucamera\ at different ambient temperatures. 
To collect these data, the \taucamera\ was placed inside an environmental chamber in front of a scientific-grade blackbody (\blackbody). 
The blackbody and environmental chamber were cycled to different pairs of $(\tamb,\tobj)$, and frames were acquired for the different permutations.
\cref{fig:nonuniformity} provides an example of the collected data.

The \taucamera\ was modeled by the image-acquisition model in \cref{eq:acquisition:frame}. 
The calibration was done according to Nugent et al.~\cite{Nugent2013}, using a third-degree polynomial to approximate the coefficients $g,d$.
For each pixel in the sensor, \cref{eq:acquisition:frame} can be formulated as:
\begin{equation}\label{eq:methods:coefficients}
    I_p(\tobj,\tamb)  = \sum^3_{i=0}\left(g_{i,p}\cdot\tamb^i\cdot{\tobj}_p^4  + d_{i,p}\cdot\tamb^i\right)
\end{equation} where $g_{i,p},d_{i,p}$ are the i-th gain and offset coefficients at pixel $p$, respectively.
Equation~\ref{eq:methods:coefficients} can be rewritten as a matrix multiplication:
\begin{equation}\label{eq:methods:coefMatMul}
    \begin{split}
        &\begin{matrix}
            T^n_p =& [{\tobj^4}_n&\hdots&{\tobj^4}_n{\tamb^3}_n&1&\hdots&{\tamb^3}_n]\\
            C_p   =& [g_{0,p}&\hdots&g_{3,p}&d_{0,p}&\hdots&d_{3,p}] ^T
        \end{matrix}\\
        &I_{N,p} \equiv  T_{N,p}\cdot C_p
    \end{split}
\end{equation} where $T^n_p$ contains the appropriate temperatures of the n-th sample of a permutation, $T_{N,p}$ is a matrix with all of the temperatures corresponding to all of the samples of the permutation as rows, and $I_{N,p}$ is a matrix with all of the acquired samples as rows.
The coefficients $C_p$ in \cref{eq:methods:coefMatMul} are found by solving the least-squares problem:
\begin{equation}\label{eq:methods:coefLS}
    C_p = T_{N,p}^\dagger\cdot I_{N,p}
\end{equation} where $T_{N,p}^\dagger$ is the Moore-Penrose pseudo inverse.

Finally, we stack all of the 2D coefficient maps $C_p$ into a 3D tensor $\mat{C}$, with $\mat{C}[0]$ being the 2D map of coefficient $g_0$ etc.

The degradation model described in \cref{eq:methods:coefLS} is per pixel, and therefore unique to each camera. 
This means that nonuniformity will also be modeled by the coefficients (e.g., dead pixels, FPN), limiting the usability of the degradation model to only the specific camera that collected the data.
To enable generalization of the degradation model for different cameras, the final stage in the model exploits the circular symmetry of the nonuniformity and uses the dependency between neighboring pixels.

Nonuniformity has a circular symmetry around the middle of the frame~\cite{IrFundamentals}. 
This is due to the ambient temperature of the camera, generating radiation from the chassis and lens, which is also reflected onto the sensor. 
Rays of thermal radiation from the body of the camera travel to the sensor and affect each pixel differently. The superposition of these rays on each pixel creates the circular symmetry of the nonuniformity.
An example of the circular symmetry can be seen in \cref{fig:nonuniformity}. 

The spatial dependency was modeled as a radial map around the middle of the frame. 
The radial map was constructed from two mesh grids $\mat{H},\mat{W}$ with the same dimensions as the frames. Each row in $\mat{H}$ and each line in $\mat{W}$ runs from $-0.5$ to $0.5$, such that $\mat{H}=\mat{W}^T$. The radial map is defined as:
\begin{equation}
    \begin{split}
        \mat{R} &= \sqrt{\mat{H}^2+\mat{W}^2},\quad\quad\mat{H},\mat{W},\mat{R}\in\mathcal{R}^{h,w}\\
        \mat{R} &= \sqrt{\begin{bmatrix} 
                            -0.5&\ldots&-0.5 \\
                            \vdots & \ddots & \vdots \\
                            0.5&\ldots&0.5 
                            \end{bmatrix}^2+\begin{bmatrix} 
                            -0.5&\ldots&0.5 \\
                            \vdots & \ddots & \vdots \\
                            -0.5&\ldots&0.5 
                            \end{bmatrix}^2}
    \end{split}
\end{equation} where $h,w$ are the dimensions of the frames. The power of the matrix is performed element-wise.

The coefficient maps are modeled as:
\begin{equation}\label{eq:methods:approxCoefR}
    \Hat{\mat{C}}[i] = \sum_{j=0}^M{m_j\cdot\mat{R}^j},\quad\quad m_j\in\mathcal{R},\ \mat{C},\mat{R}\in\mathcal{R}^{h,w}
\end{equation} where $m_j$ is the spatial coefficient, and $M$ is the number of spatial coefficients.
Least-squares is solved to find the spatial coefficients.

A frame from a given temperature map is estimated by:
\begin{equation}\label{eq:methods:estimateFrame}
    \Hat{I}(\tobj,\tamb)=T(\tobj,\tamb)\cdot\Hat{C}
\end{equation} where $T(\tobj,\tamb)$ is the temperature vector in \cref{eq:methods:coefMatMul}.

The final degradation model was noiseless and only contained low frequencies. Random FPN and Gaussian noise were added to the model during training. This enabled the network to converge to a general solution that is applicable on different cameras with various degradation profiles.

\subsection{Training procedure}\label{sec:methods:training}
    \noindent The network was written in Python~3.10~\cite{python} using Pytorch~1.13~\cite{pytorch}, and was trained on a single Nvidia Titan A100. 
The seed was set to $42$, and the CUDNN backend was set to deterministic mode.
The network was trained using the ADAM optimizer \cite{adamOpt2015} with a learning rate of $10^{-4}$. The learning rate was halved on a validation loss plateau of more than 3 epochs. The network was run for 60 epochs with batches of 16, meaning that each epoch was roughly 800 iterations. Early stopping was applied for a validation loss plateau of 8 epochs. 
The weights were initialized using the orthogonal scheme \cite{orthoInit2014} with a scaling of $10^{-1}$.
The hyperparameter search was run twice with different seeds (42 and 24), and the best results were chosen.

Multiframes were simulated by randomly sampling homographies for each frame in the dataset, creating different views of the same frame.
The inverse homographies were used to register all of the views toward the original frame, which was set as the pivot frame $\mathcal{I}$. 
The sampled homographies either created a random walk from one side of the temperature map to the other, or a hover above a random point in the frame. 
The overlaps between the different views were randomly set between $60\%$ and $80\%$, similar to a UAV flight scenario, as seen in \cref{sec:results:realdata}.
Homography and frame warping was implemented with the package Kornia~v0.67.

Imperfect registration was simulated by randomly adding perturbations to the inverse homographies: random translation of up to $\pm2$ pixels and noise from the distribution $\mathcal{N}(0, 5\cdot10^-5)$ to the perspective elements of the homography (commonly known as $h_{31}, h_{32}$).
Random horizontal and vertical flips, and $90^\circ$ rotations were applied to the frames before the homography sampling.

The gray-level frames were cropped to $128\times 128$ patches before entering the network. For validation, a constant cropping was applied around the middle of the frame, and no other augmentations were applied.

Random Gaussian noise with $\sigma^2=5$ gray levels and FPN were generated for each frame (\cref{sec:methods:data}). FPN was generated as:
\begin{equation}\label{eq:methods:fpn}
    \begin{bmatrix}
        1 \\ \vdots  \\ 1
    \end{bmatrix}_{h\times1}\cdot
    \begin{bmatrix}
        U[u_{\min}, u_{\max}] \\ \vdots  \\ U[u_{\min}, u_{\max}]
    \end{bmatrix}^T_{1\times w}
\end{equation} where $U$ is uniform distribution. $u_{\min}, u_{\max}$ were chosen as $u_{\min}=0.9, u_{\max}=1.01$. 
The Gaussian noise and FPN were only generated once for each frame and used throughout the entire validation process for reproducibility of results between experiments.

Normalization to range [0,1] was applied to both the temperature map and the gray-level frame.
Throughout the training and validation sets, the maximal and minimal values of the temperature maps and the maximal and minimal values of the gray-level frames were obtained.
The normalization was applied on the temperature maps as:
\begin{equation}\label{eq:methods:norm_t}
    \Bar{X} = \frac{X - X_\text{min}}{X_\text{max}-X_\text{min}}
\end{equation}
where $\Bar{X}$ is the normalized input and $X_\text{min}, X_\text{max}$ are the minimal and maximal temperatures, respectively, over all datasets.
Normalization for the gray-level frames was applied as:
\begin{equation}\label{eq:methods:norm_frame}
    \Bar{I}(\tamb) = \frac{I(\tamb) - I_\text{min}}{I_\text{max}-I_\text{min}}
\end{equation}
where $\Bar{I}$ is the normalized gray-level frame and $I_\text{min},I_\text{max}$ are the minimal and maximal gray levels, respectively, over all datasets.

The following pipeline summarizes the creation of samples for the network.
First, an accurate temperature map is sampled from the dataset. $\nFrames$ homographies are randomly sampled and applied to the temperature map to create an overlapping burst of frames.
The model described in \cref{sec:methods:synthData} is applied to each frame in the burst to turn it into a gray-level frame \cref{eq:methods:estimateFrame}.
The same FPN is applied to all frames in the burst \cref{eq:methods:fpn}, and random noise is applied to each frame in the burst separately.
Finally, normalization is applied to the ambient temperature \cref{eq:methods:norm_t} and overlapping gray-level frames \cref{eq:methods:norm_frame}, and both are passed to the network.
\subsection{Data}\label{sec:methods:data}
    \noindent The dataset used for training consisted of $12,897$ frames, and the validation set was composed of $4,723$ frames, all of which were captured by an UAV flying at a height of $70-100_m$ above various agricultural fields in Israel. Only clear and in-focus frames were selected for the dataset manually by a human user.

The noise variance was established by analyzing the measurements taken in the environmental chamber. All frames were stacked depth-wise, and the variance of each pixel was calculated, resulting in a 2D variance map. 
The mean of the variance map was $~5$ gray levels, which was used as the noise variance in the network training. 
The influence of $\tamb$ and $\tobj$ on $\sigma^2$ was determined to be insignificant.

To prevent data leakage between the training and validation sets and evaluate the network's ability to generalize to new data, the validation sets were captured at the same locations as the training sets but on different days. This validation approach was maintained across all training schemes to ensure a fair comparison between different experiments. The same split between the training and validation datasets was maintained throughout the study.
\section{Results}\label{sec:results}
    \begin{figure}
    \centering
    \includegraphics[width=\linewidth]{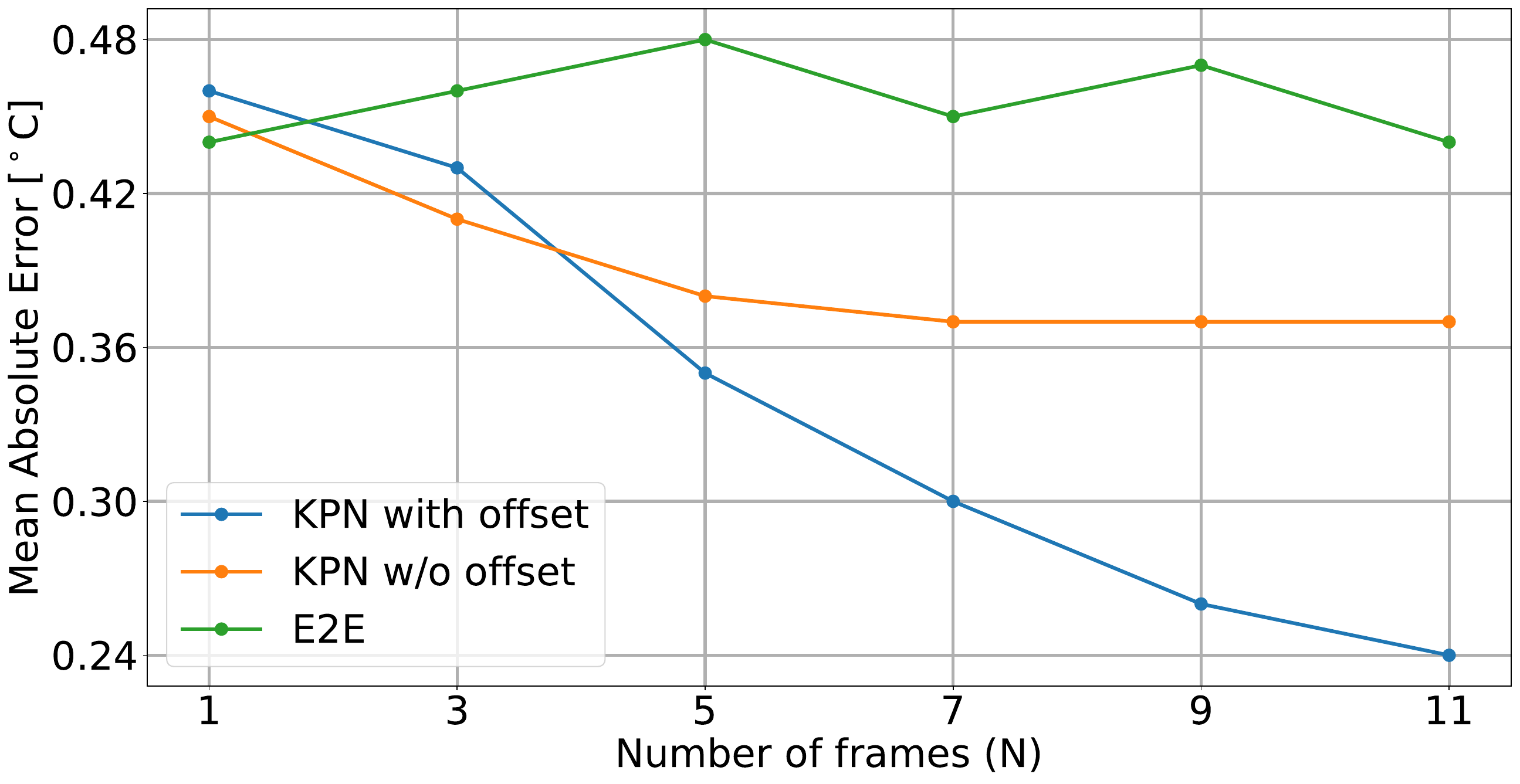}
    \caption{Mean absolute error (MAE) in degrees Celsius as a function of the number of frames $\nFrames$ for different network configurations.}
    \label{fig:metrics}
\end{figure}%
\newcommand{\heightDiffRes}{13ex}
\newcommand{\makeDiffRow}[1]{\subfloat{\includegraphics[height=\heightDiffRes]{res_diff/ART_#1/rect.pdf}}
    \hfill
    \subfloat{\includegraphics[height=\heightDiffRes]{res_diff/ART_#1/diff/7.pdf}}
    \hfill
    \subfloat{\includegraphics[height=\heightDiffRes]{res_diff/ART_#1/diff/9.pdf}}
    \hfill
    \subfloat{\includegraphics[height=\heightDiffRes]{res_diff/ART_#1/diff/11.pdf}}
    \hfill
    \subfloat{\includegraphics[height=\heightDiffRes]{res_diff/ART_#1/cbar.pdf}}}
\newcommand{\makeDiffFig}[7]{\begin{figure}
        \centering
        \makeDiffRow{#2}
        \\
        \makeDiffRow{#3}
        \\
        \makeDiffRow{#4}
        \\
        \makeDiffRow{#5}
        \\
        \makeDiffRow{#6}
        \\
        \makeDiffRow{#7}
        \\
        \setcounter{subfigure}{0}
        \subfloat[GT]{\includegraphics[height=\heightDiffRes]{res_diff/ART_#1/rect.pdf}}
        \hfill
        \subfloat[7]{\includegraphics[height=\heightDiffRes]{res_diff/ART_#1/diff/7.pdf}}
        \hfill
        \subfloat[9]{\includegraphics[height=\heightDiffRes]{res_diff/ART_#1/diff/9.pdf}}
        \hfill
        \subfloat[11]{\includegraphics[height=\heightDiffRes]{res_diff/ART_#1/diff/11.pdf}}
        \hfill
        \subfloat{\includegraphics[height=\heightDiffRes]{res_diff/ART_#1/cbar.pdf}}
        \caption{Difference between our temperature estimation and the GT\@.
        The left-most figure is the GT\@.
        The next figures are zoom-ins of the areas inside the red rectangles.
        The number of frames used for the estimation is below each map, from left to right 7, 9 and 11 frames. 
        The mean absolute error between the GT and our estimation is written in the top-left corner of each map.}
        \label{fig:results:diff}
    \end{figure}}
\makeDiffFig{NeveYaar_210520_1}{NeveYaar_210520_2}{MevoBytar_210818_2}{Tzora_210523_9}{NirEliyho_211005_0}{NirEliyho_211005_10}{MevoBytar_210818_1}
\newcommand{\heightPatchRes}{0.16\linewidth}%
\newcommand{\makePatchRow}[1]{\subfloat{\includegraphics[height=\heightPatchRes]{res_patch/#1/sample.pdf}}
    \hfill
    \subfloat{\includegraphics[height=\heightPatchRes]{res_patch/#1/label.pdf}}
    \hfill
    \subfloat{\includegraphics[height=\heightPatchRes]{res_patch/#1/Ours.pdf}}
    \hfill
    \subfloat{\includegraphics[height=\heightPatchRes]{res_patch/#1/ADMIRE.pdf}}
    \hfill
    \subfloat{\includegraphics[height=\heightPatchRes]{res_patch/#1/DeepIR.pdf}}
    \hfill
    \subfloat{\includegraphics[height=\heightPatchRes]{res_patch/#1/He_et_al.pdf}}
}%
\newcommand{\makePatchesFig}[5]{\begin{figure*}
        \centering
        \makePatchRow{#1}\\
        \makePatchRow{#2}\\
        \makePatchRow{#3}\\
        \makePatchRow{#4}\\
        \setcounter{subfigure}{0}
        \subfloat[Sample]{\includegraphics[height=\heightPatchRes]{res_patch/#5/sample.pdf}}
        \hfill
        \subfloat[GT]{\includegraphics[height=\heightPatchRes]{res_patch/#5/label.pdf}}
        \hfill
        \subfloat[This study]{\includegraphics[height=\heightPatchRes]{res_patch/#5/Ours.pdf}}
        \hfill
        \subfloat[ADMIRE~\cite{Tendero12}]{\includegraphics[height=\heightPatchRes]{res_patch/#5/ADMIRE.pdf}}
        \hfill
        \subfloat[DeepIR~\cite{Saragadam2021}]{\includegraphics[height=\heightPatchRes]{res_patch/#5/DeepIR.pdf}}
        \hfill
        \subfloat[He et al.~\cite{He2018}]{\includegraphics[height=\heightPatchRes]{res_patch/#5/He_et_al.pdf}}
        \caption{Zoomed-in results of the different methods.
        The left-most figure is the reference frame with a red rectangle. 
        The following figures in each row are the results of the areas inside the red rectangles.
        Number of frames $\nFrames=11$ for all results.}
        \label{fig:results:patches}
    \end{figure*}}%
\makePatchesFig{180725_Ramon_1}{180805_Peach_31}{180805_Peach_37}{MevoBytar_210818_69}{YanivReshef_190816_32}%
\begin{figure*}
    \centering
    \subfloat[KPN with offset]{\includegraphics[width=0.3\linewidth]{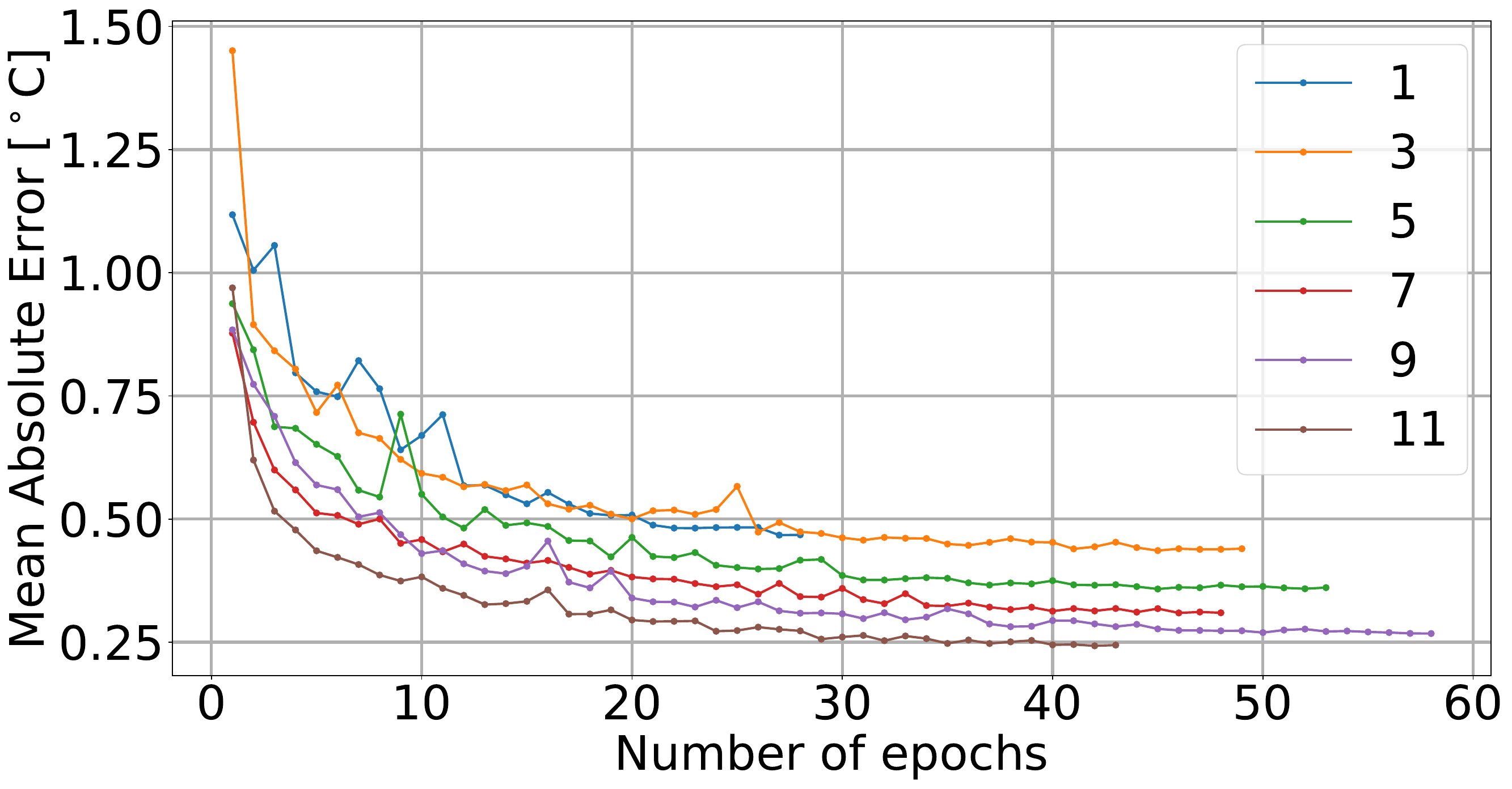}}\hfill
    \subfloat[KPN w/o offset]{\includegraphics[width=0.3\linewidth]{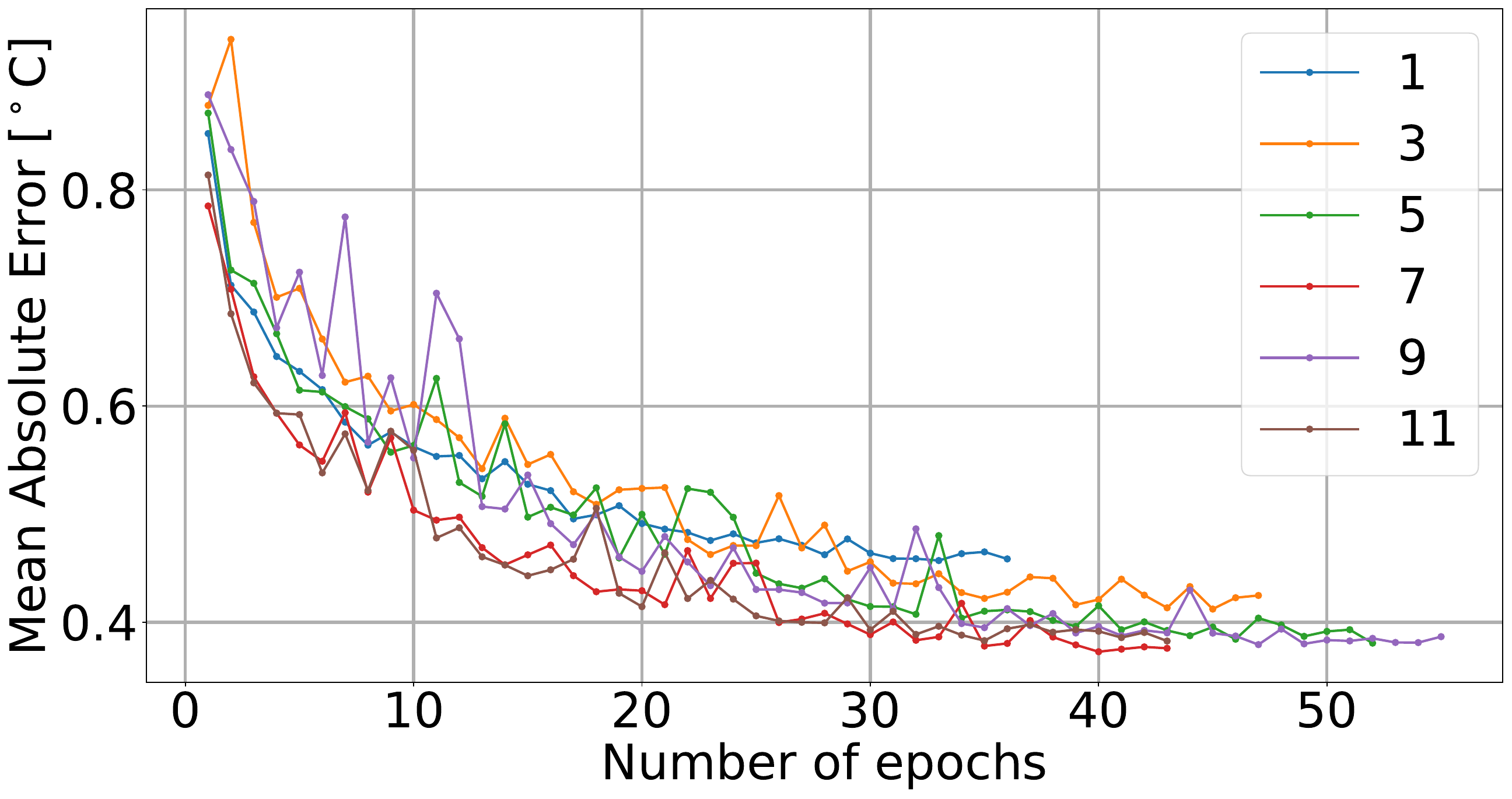}}\hfill
    \subfloat[E2E]{\includegraphics[width=0.3\linewidth]{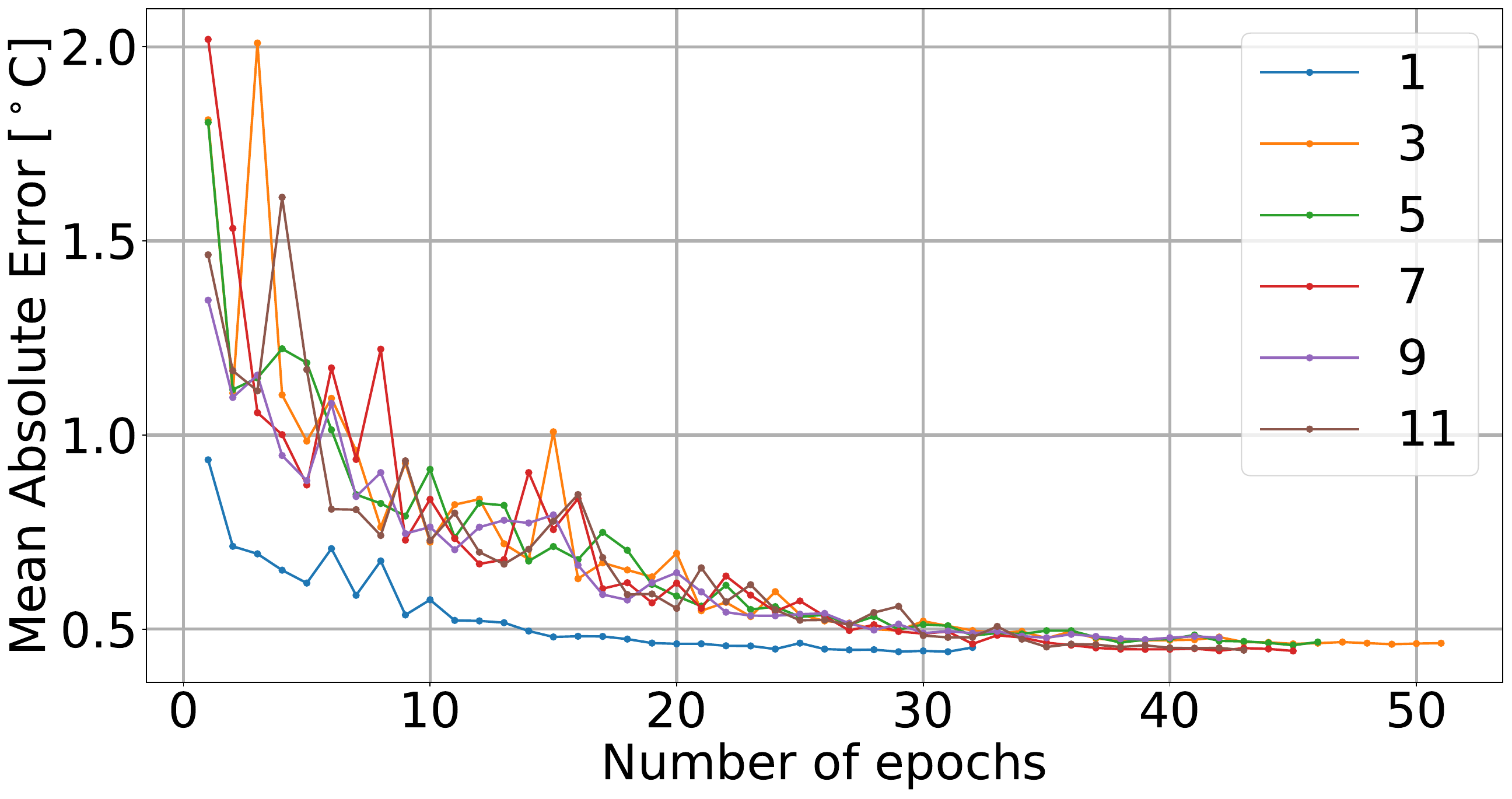}}
    \caption{Convergence of the validation mean absolute error (MAE) loss for different architectures: (a) kernel prediction network (KPN) with offset. (b) KPN without offset. (c) End to end (E2E). Each color represents a different number of frames $\nFrames$.}
    \label{fig:results:convergence}
\end{figure*}%
\noindent To demonstrates the efficacy of our method, we compared the mean absolute error (MAE) of the temperature estimation with different blocks of the network, namely with and without the offset estimation block, and using an end-to-end (E2E) network instead of the KPN architecture. The results are displayed as a function of the number of frames $\nFrames$ in \cref{fig:metrics}.
Whereas our method can handle misalignments between frames, other methods require a perfect alignment, which is impossible in real-world scenarios.
As a result, comparing temperature estimation with other methods is impossible, and we could only compare the NUC between our method and other methods on pivot frame $\mathcal{I}$. Moreover, the other methods are not radiometric, and thus cannot be used for temperature estimation.

\cref{fig:metrics} demonstrates the superiority of our method, evidence by the low MAE for almost every number of frames.
The E2E was a UNET~\cite{unet} architecture similar to our KPN network.
The main difference was that the last layer estimated the per-pixel result instead of outputting the kernel.
A hyperparameter search was also performed for the E2E solution for fair comparison (number of channels and normalization).
The MAE of the E2E network in \cref{fig:metrics} was unaffected by the number of frames, whereas the KPN results improved with the number of frames, indicating that the E2E network only uses the reference frame. Moreover, the MAE results for E2E were worse than for the KPN network.

The offset block greatly improved the results with increasing number of frames as seen in \cref{fig:metrics}, with an over $0.1^\circ C$ improvement for KPN and more than $0.2^\circ C$ improvement relative to E2E for $\nFrames=11$, indicating that the offset block is beneficial. Results suggested that increasing the number of frames without the offset block reaches a plateau at around $\nFrames=5$ and does not improve the results further, in contrast to the offset block which continues to improve the results with more frames. Since the offset block is lightweight, it offers significant improvement with little computational cost.

The effect of the number of frames $\nFrames$ is shown in \cref{fig:results:diff}. The figure shows per-pixel error in the temperature estimation for different numbers of frames.
The left-most figure (column a) shows the GT temperature map. 
The absolute per-pixel difference between the GT and our method's estimation for the area inside the red rectangle is shown in the next columns (b, c, and d). 
Each figure shows the estimated temperature for a different number of frames: (b) $N=7$ frames, (c) $N=9$ frames and (d) $N=11$ frames. 
The color bar on the right of each row shows the error range for the row in degrees Celsius. 
The MAE $^\circ C$ between the GT and the estimation is written in the top-left corner of each difference map. 
Each row is a different frame.
The improvement caused by the number of frames is clearly reflected by the homogeneity in the difference map and the decreasing MAE as a function of $\nFrames$.
More examples are available in the supplementary material, \cref{supp_diff_7,supp_diff_8,supp_diff_9,supp_diff_10,supp_diff_11}.

Because other methods are not radiometric and essentially only improve the appearance of a frame, we could only compare NUC results with other methods. 
\cref{fig:results:patches} displays the NUC results of different methods. Column (a) shows the reference sample frame. Column (b) shows the GT temperature map. Column (c) shows the results of our method. Column (d) shows the results of ADMIRE~\cite{Tendero12} performed on each frame separately and then registered and averaged. Column (e) shows the estimation of DeepIR~\cite{Saragadam2021} and column (f) shows the estimation of He et al.~\cite{He2018}.
All results were obtained with $\nFrames=11$.

As evidenced by \cref{fig:results:patches}, our NUC method was better than the others.
ADMIRE~\cite{Tendero12} failed to rectify the FPN,
DeepIR~\cite{Saragadam2021} hallucinated details (e.g., the deformation in the junction in the fourth row, or the abrupt black to white edge in the fifth row). 
He et al.'s~\cite{He2018} method failed to handle the FPN\@.
Both DeepIR~\cite{Saragadam2021} and He et al.~\cite{He2018} methods oversmoothed the results.
These methods have low fidelity, and thus are unable to serve for temperature estimation.
More results are available in the supplementary material, \cref{supp_patch_1,supp_patch_2,supp_patch_3,supp_patch_4,supp_patch_5,supp_patch_6}.

\cref{fig:results:convergence} depicts the convergence of the validation MAE loss of the E2E and of the KPN with and without the offset block, as a function of the number of frames. 
Notice that the loss for the E2E networks converges to roughly the same value for all numbers of frames, whereas the KPN-based networks give different results as a function of the number of frames $\nFrames$.
When comparing the convergence with and without the offset block, it seems that the offset block has a smoothing effect on the validation loss. This effect might occur because the KPN can concentrate on correcting the NUC, while the offset block handles the temperature estimation.

\subsection{Real data}\label{sec:results:realdata}
    \newcommand{\sizeRealData}{0.382}%
\begin{figure*}
    \centering
    \subfloat[]{\includegraphics[width=\sizeRealData\linewidth]{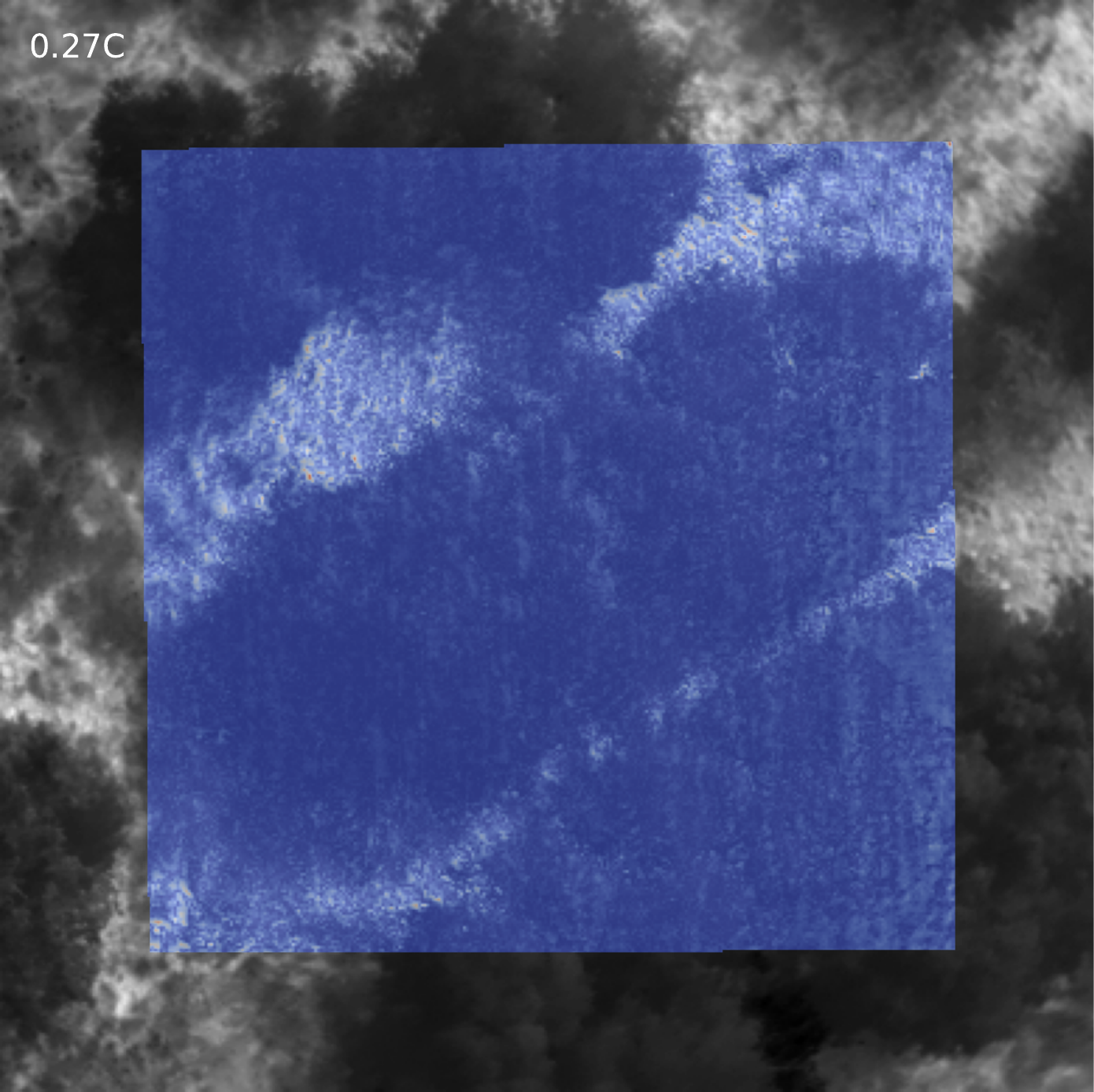}\label{fig:results:realdata:a}}
    \hfill
    \subfloat[]{\includegraphics[width=\sizeRealData\linewidth]{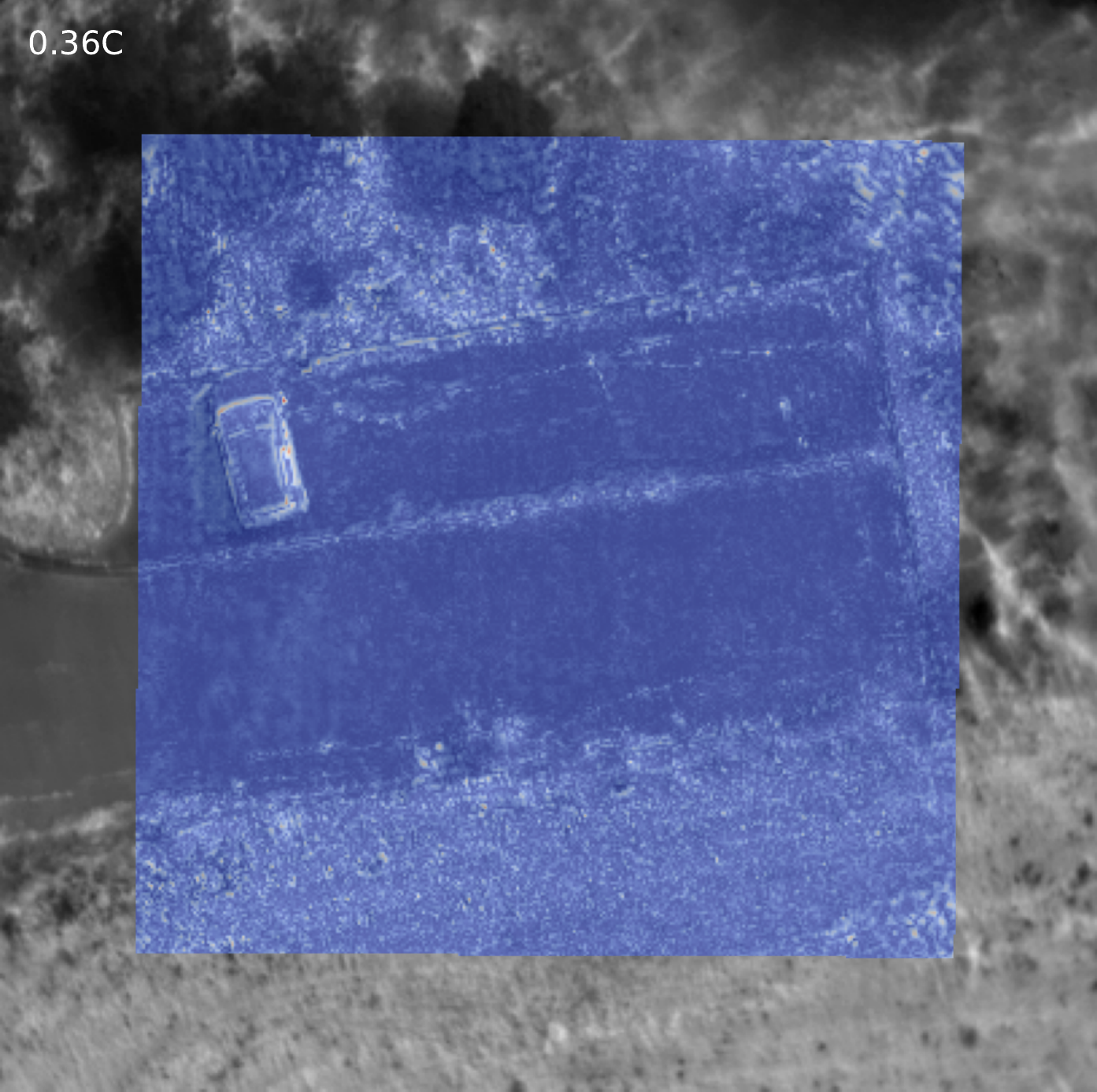}\label{fig:results:realdata:b}}
    \\
    \subfloat[]{\includegraphics[width=\sizeRealData\linewidth]{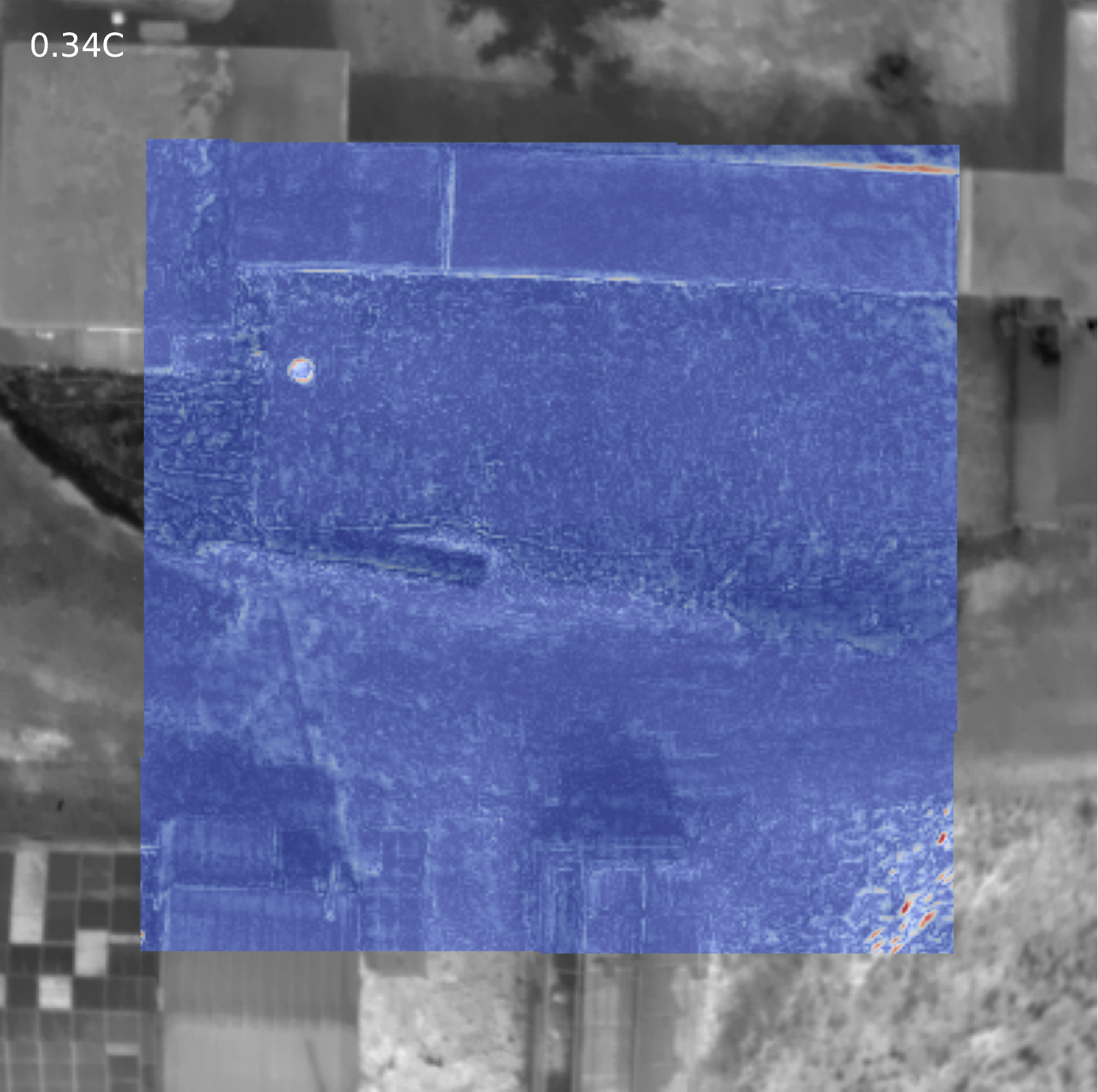}\label{fig:results:realdata:c}}
    \hfill
    \subfloat[]{\includegraphics[width=\sizeRealData\linewidth]{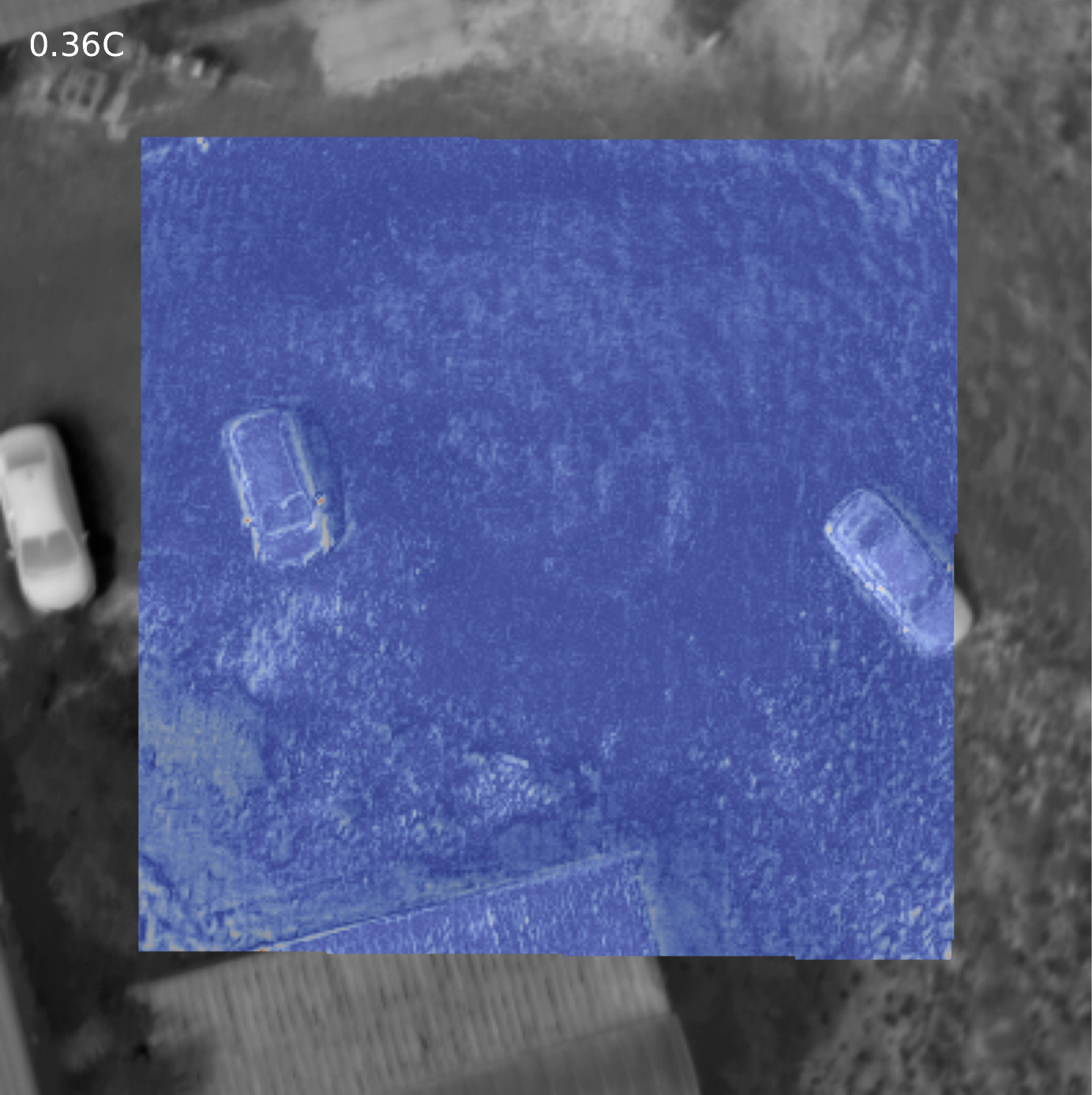}}
    \\
    \subfloat[]{\includegraphics[width=\sizeRealData\linewidth]{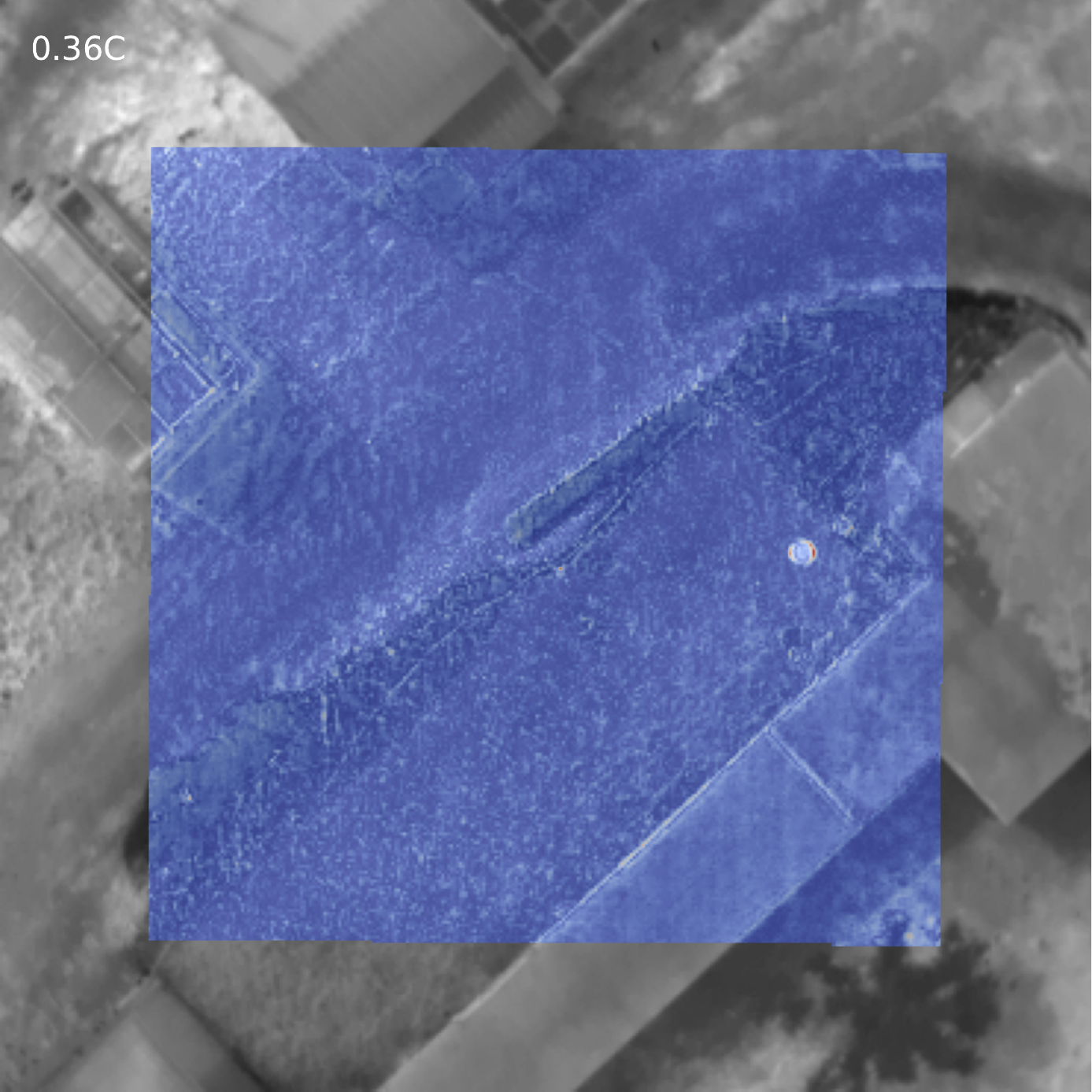}}
    \hfill
    \subfloat[]{\includegraphics[width=\sizeRealData\linewidth]{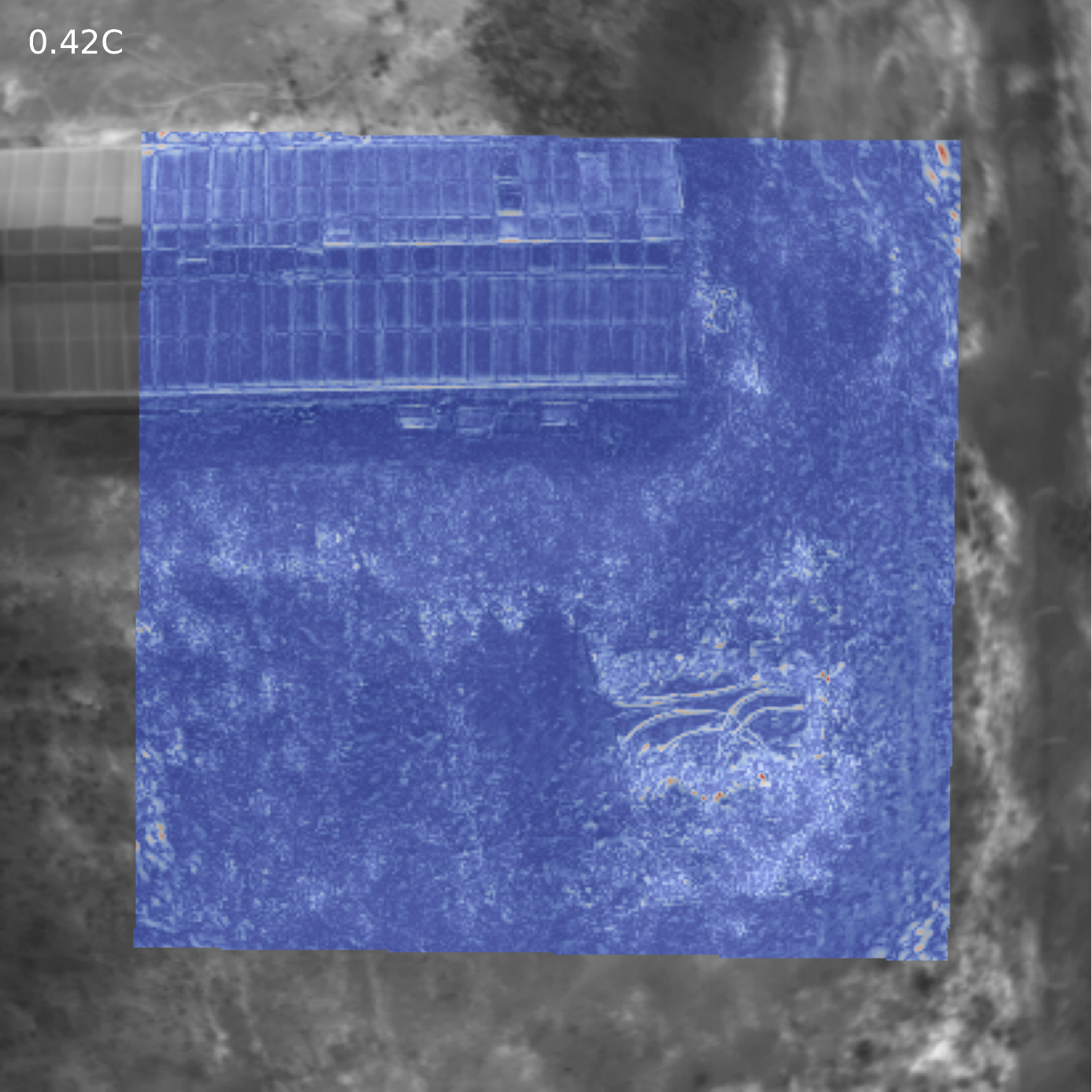}}
    \caption{Various results on real data. The gray background is the ground truth (GT) temperature map and the colored map is the difference between the GT and estimated temperature maps.
    The number in the top-left corner is the mean absolute error (MAE) between the estimated and GT temperature maps.
    }
    \label{fig:results:realdata}
\end{figure*}%
\begin{figure*}
    \centering
    \subfloat[]{\includegraphics[width=.3\linewidth]{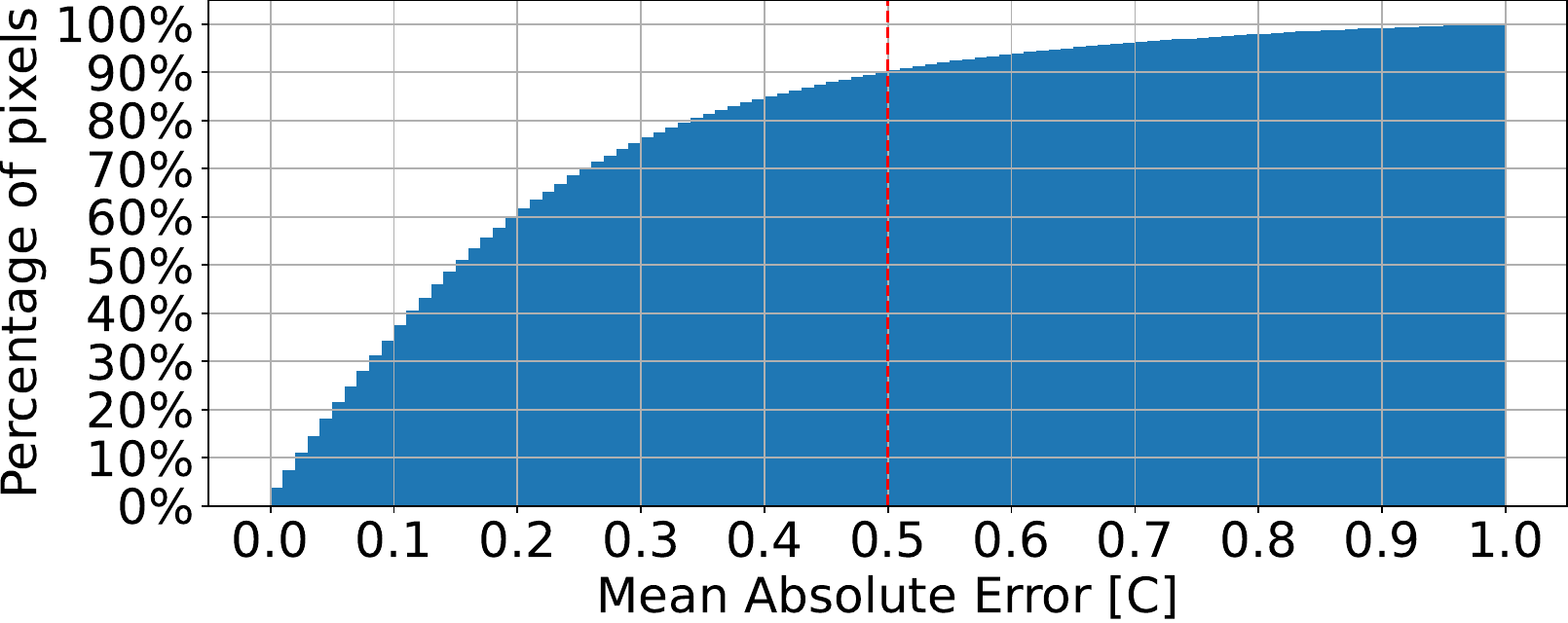}}
    \subfloat[]{\includegraphics[width=.3\linewidth]{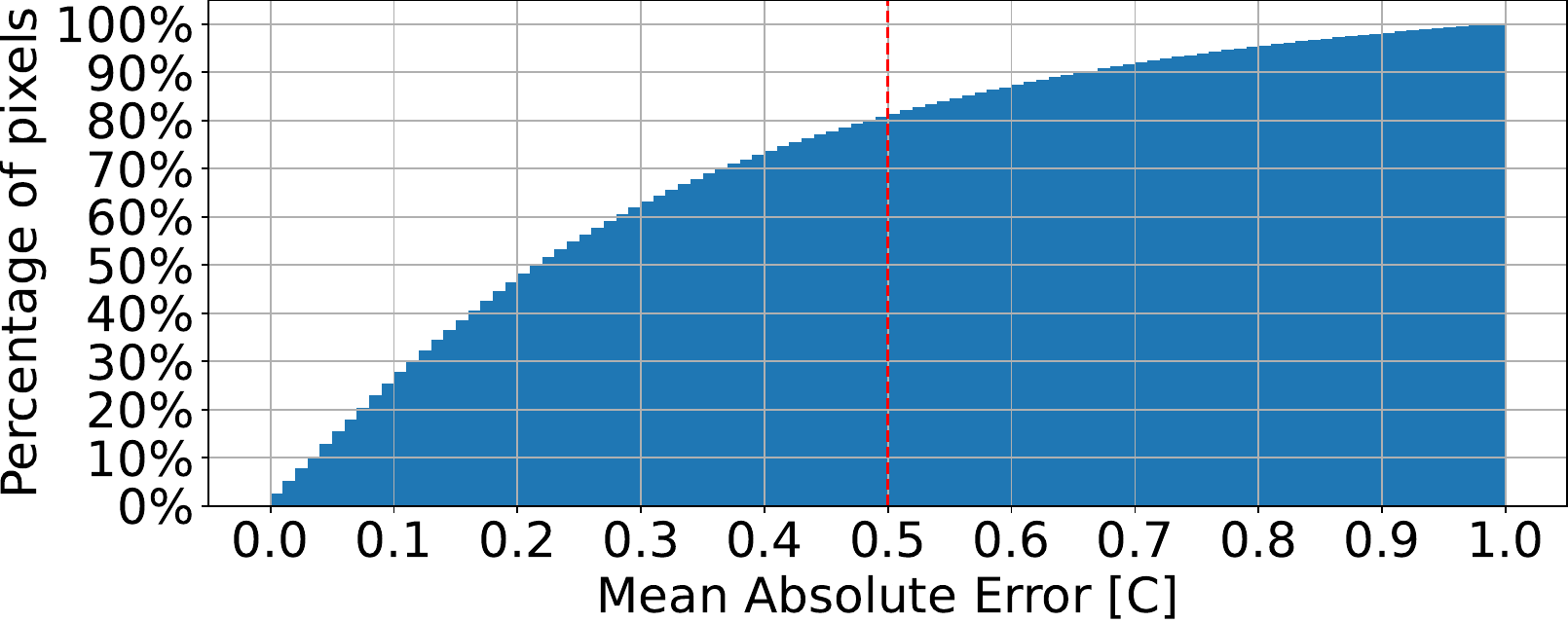}}
    \subfloat[]{\includegraphics[width=.3\linewidth]{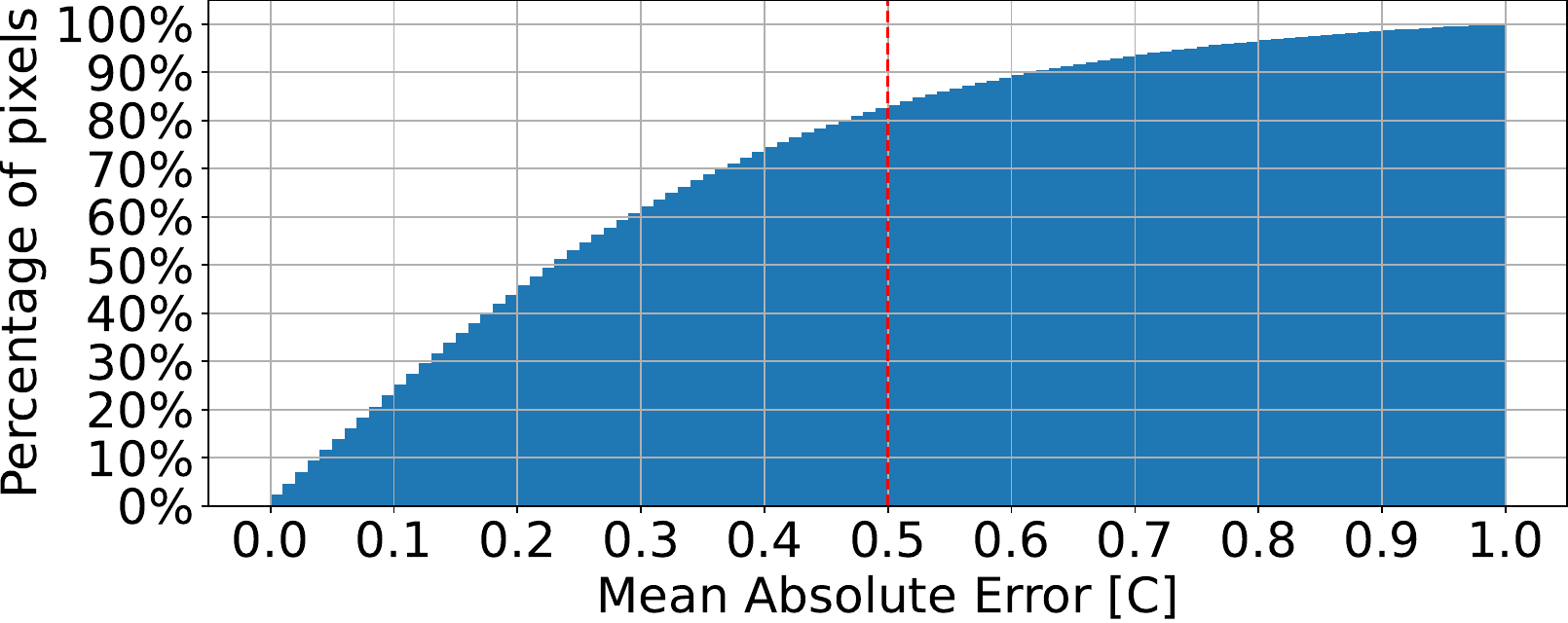}}
    \caption{Mean absolute error (MAE) of pixels as a cumulative function for real data. The y-axis is the percentage of pixels with MAE less than the value on the x-axis.
    Panel (a) is the MAE of \cref{fig:results:realdata:a}. Panel (b) is the MAE of \cref{fig:results:realdata:b}. Panel (c) is the MAE of \cref{fig:results:realdata:c}.}
    \label{fig:results:realdata:percentage}
\end{figure*}%
We validated the effectiveness of the proposed method on real data. Two cameras, \scientificCamera\ and \taucamera, were attached to a \hyperlink{https://www.dji.com/global/matrice600}{DJI Matrice 600} UAV and both captured the same scenes in nadir view at a height of $50_m$ above the ground at a vertical speed of $10_{m/s}$.
The \scientificCamera\ is a scientific-grade radiometric camera which outputs a temperature map of the scene, whereas the \taucamera\ outputs a gray-level map corresponding to the radiation flux. An image of the setup can be found in \cref{supp_uav} of the supplementary material.
Notice that the \taucamera\ used for the experiment was not the one used to collect the calibration data in \cref{sec:method}, which further strengthens the generality and robustness of the proposed method.

The frame rate for the \taucamera\ was set to $30_{Hz}$; the resolution of the \taucamera\ was $336\times256$ pixels, the focal length was $9.8_{mm}$, and the sensor size was $4.4_{mm}$ per 256 pixels (in the direction of the flight). The ground sampling distance was $\frac{50\cdot4.4}{9.8 \cdot 256}=0.087_{m/pix}$. The drone passed $\frac{10}{30}=0.33_m$ for every frame. This means that an object moved $\frac{0.33}{0.087}=3.80_{pix}$ between consecutive frames. Thus, an object could appear in $\frac{256}{3.80}\cong67$ frames. The \scientificCamera\ field of view was much larger than that of the \taucamera\, so a frame of the \scientificCamera\ contained multiple frames of the \taucamera.

The \scientificCamera\ requires accurate ambient parameters to produce a valid temperature map.
The ambient temperature and humidity were gathered from a nearby weather station ($28.4^\circ C$ and $32\%$, respectively). The emissivity was tuned using an accurate temperature sensor placed in the scene.

Both cameras were focused to infinity. 
The flight height of $50_m$ above ground ensured that all objects where within the depth of field of both cameras.
\scientificCamera\ captured $1,192$ frames at $5_{Hz}$ and the \taucamera\ captured $7,152$ frames at $30_{Hz}$.

The frames of the \taucamera\ were divided into overlapping groups of 7 frames each. We used 7 frames due to hardware limitations.
The frames of each group were registered toward the middle frame of the group.
The registration was performed by SIFT feature-matching using the Python package Kornia V0.6.10.
The registered frame groups were the input to the network.

The output of the network was the estimated temperature map of the scene. These estimated temperature maps were registered to the \scientificCamera\ temperature maps by hand-picking correspondence points. The final registration was performed using the Python package OpenCV V4.5.1.
The GT and estimated temperature maps are presented in \cref{supp_H,supp_I,supp_A,supp_O,supp_B,supp_M} in the supplementary material.

Six results are presented in \cref{fig:results:realdata} and four more are presented in \cref{supp_realdata} in the supplementary material.
We present the difference maps between the estimated and GT temperature maps, produced by the proposed method, in each subfigure. The GT maps are in gray, and the color scale from blue to red is the magnitude of the difference, with blue denoting low and red denoting high errors. The upper-left corner of the upper image displays the MAE of the difference map as a white number.

The MAE values span $0.27-0.54^\circ C$, indicating a high accuracy of temperature estimation comparable to the \scientificCamera\ precision of ($\sim0.5^\circ C$). This was obtained without applying any thermographic corrections or NUC to the \taucamera\ data, relying solely on the raw measurements of the radiation flux as gray levels. The supplementary material provides the detailed configuration of the \taucamera.

The cumulative distribution function of the MAE between the GT temperature map and the estimated temperature map is shown in \cref{fig:results:realdata:percentage}.
The dashed red lines indicate the $0.5^\circ C$ threshold. All three examples show that more than $80\%$ of the pixels have a MAE of less than $0.5^\circ C$.
This further solidifies the effectiveness of the proposed method.
\section{Conclusion}\label{sec:conclusion}
    \noindent We presented a novel method for simultaneous temperature estimation and NUC in \ir imaging, based on a DL method that incorporates the physical model of the sensor. 
The method uses redundant information between multiple overlapping frames to infer the scene temperature and correct nonuniformity, without requiring any calibration or external reference. 
The method also exploits prior knowledge of the camera's ambient temperature, which is measured by a built-in sensor, to improve the accuracy and robustness of the estimation.

We evaluated the performance of the method on synthetic and real data and compared it with existing methods. 
The results showed that the method can achieve high accuracy and low error, and can handle various scenarios, such as changing ambient temperature, moving objects, and complex backgrounds.

We showed that performance improves with the number of frames, highlighting the benefits of exploiting the redundant information between frames.
The training process introduced misalignments between frames, which were handled by the method and did not affect its performance.
The method can also generalize well to different camera models and settings, and can be easily adapted to different applications. This was demonstrated by real data collected with a different camera mounted on an UAV\@. The MAE with the real UAV data was $0.27-0.54^\circ C$, which is comparable to the accuracy of scientific-grade cameras.

The method offers a simple and effective solution for improving the quality and reliability of low-cost uncooled \ir imaging and can potentially enable new applications that require accurate and consistent temperature measurements.
\section*{Acknowledgments}
The authors are deeply grateful for the help of Ohaliav Keisar with the UAV data collection.

The authors thank Dr. Yaffit Cohen and Eitan Goldstein for the UAV data used in this work; 
and Moti Barak, Lavi Rosenfeld and Liad Reshef for the design and construction of the environmental chamber.

\section*{Funding}
The research was funded by the Israeli Ministry of Agriculture's Kandel Program under grant no. 20-12-0030.

\section*{Disclosures}
\noindent The authors declare no conflicts of interest.

\balance    
\bibliographystyle{IEEEtran}
\bibliography{biblography}

\begin{thebibliography}{10}
\providecommand{\url}[1]{#1}
\csname url@samestyle\endcsname
\providecommand{\newblock}{\relax}
\providecommand{\bibinfo}[2]{#2}
\providecommand{\BIBentrySTDinterwordspacing}{\spaceskip=0pt\relax}
\providecommand{\BIBentryALTinterwordstretchfactor}{4}
\providecommand{\BIBentryALTinterwordspacing}{\spaceskip=\fontdimen2\font plus
\BIBentryALTinterwordstretchfactor\fontdimen3\font minus
  \fontdimen4\font\relax}
\providecommand{\BIBforeignlanguage}[2]{{%
\expandafter\ifx\csname l@#1\endcsname\relax
\typeout{** WARNING: IEEEtran.bst: No hyphenation pattern has been}%
\typeout{** loaded for the language `#1'. Using the pattern for}%
\typeout{** the default language instead.}%
\else
\language=\csname l@#1\endcsname
\fi
#2}}
\providecommand{\BIBdecl}{\relax}
\BIBdecl

\bibitem{ir_importance_1}
\BIBentryALTinterwordspacing
H.~D. Adams, M.~Guardiola-Claramonte, G.~A. Barron-Gafford, J.~C. Villegas,
  D.~D. Breshears, C.~B. Zou, P.~A. Troch, and T.~E. Huxman, ``Temperature
  sensitivity of drought-induced tree mortality portends increased regional
  die-off under global-change-type drought,'' \emph{Proceedings of the National
  Academy of Sciences}, vol. 106, no.~17, pp. 7063--7066, 2009. [Online].
  Available: \url{https://www.pnas.org/doi/abs/10.1073/pnas.0901438106}
\BIBentrySTDinterwordspacing

\bibitem{ir_importance_2}
\BIBentryALTinterwordspacing
H.~G. Jones, ``{Monitoring plant and soil water status: established and novel
  methods revisited and their relevance to studies of drought tolerance},''
  \emph{Journal of Experimental Botany}, vol.~58, no.~2, pp. 119--130, 09 2006.
  [Online]. Available: \url{https://doi.org/10.1093/jxb/erl118}
\BIBentrySTDinterwordspacing

\bibitem{bolometer}
R.~Bhan, R.~Saxena, C.~Jalwania, and S.~Lomash, ``Uncooled infrared
  microbolometer arrays and their characterisation techniques,'' \emph{Defence
  Science Journal}, vol.~59, p. 580, 11 2009.

\bibitem{IrFundamentals}
M.~Vollmer and K.-P. Mllmann, \emph{Infrared Thermal Imaging: Fundamentals,
  Research and Applications}, 2nd~ed.\hskip 1em plus 0.5em minus 0.4em\relax
  Wiley-VCH, 2018.

\bibitem{Riou2004}
O.~Riou, S.~Berrebi, and P.~Bremond, ``Non uniformity correction and thermal
  drift compensation of thermal infrared camera,'' \emph{Thermosense XXVI},
  vol. 5405, p. 294, 2004.

\bibitem{Schulz1995}
M.~Schulz and L.~Caldwell, ``Nonuniformity correction and correctability of
  infrared focal plane arrays,'' \emph{Infrared Phys. Technol}, vol.~36, pp.
  763--777, 1995.

\bibitem{Nugent2013}
P.~W. Nugent, J.~A. Shaw, and N.~J. Pust, ``Correcting for focal-plane-array
  temperature dependence in microbolometer infrared cameras lacking thermal
  stabilization,'' \emph{Optical Engineering}, vol.~52, p. 061304, 2013.

\bibitem{Liang2017}
K.~Liang, C.~Yang, L.~Peng, and B.~Zhou, ``Nonuniformity correction based on
  focal plane array temperature in uncooled long-wave infrared cameras without
  a shutter,'' \emph{Applied Optics}, vol.~56, p. 884, 2 2017.

\bibitem{Chang2019}
S.~Chang and Z.~Li, ``Calibration algorithm for cooled mid-infrared systems
  considering the influences of ambient temperature and integration time,''
  \emph{Applied Optics}, vol.~58, p. 8118, 10 2019.

\bibitem{Oz2020}
N.~Oz, N.~Sochen, O.~Markovich, Z.~Halamish, L.~Shpialter-Karol, and I.~Klapp,
  ``Rapid super resolution for infrared imagery,'' \emph{Optics Express},
  vol.~28, p. 27196, 2020.

\bibitem{Shocher2017}
\BIBentryALTinterwordspacing
A.~Shocher, N.~Cohen, and M.~Irani, ``Zero-shot super-resolution using deep
  internal learning,'' in \emph{IEEE Conference on Computer Vision and Pattern
  Recognition (CVPR)}.\hskip 1em plus 0.5em minus 0.4em\relax IEEE, 12 2017,
  pp. 3118--3126. [Online]. Available:
  \url{https://ieeexplore.ieee.org/document/8578427}
\BIBentrySTDinterwordspacing

\bibitem{Shaham2019}
T.~R. Shaham, T.~Dekel, and T.~Michaeli, ``Singan: Learning a generative model
  from a single natural image,'' \emph{Proceedings of the IEEE International
  Conference on Computer Vision}, vol. 2019-Octob, pp. 4569--4579, 2019.

\bibitem{Scribner91}
\BIBentryALTinterwordspacing
D.~A. Scribner, K.~A. Sarkady, M.~R. Kruer, J.~T. Caulfield, J.~D. Hunt, and
  C.~Herman, ``{Adaptive nonuniformity correction for IR focal-plane arrays
  using neural networks},'' in \emph{Infrared Sensors: Detectors, Electronics,
  and Signal Processing}, T.~S.~J. Jayadev, Ed., vol. 1541, International
  Society for Optics and Photonics.\hskip 1em plus 0.5em minus 0.4em\relax
  SPIE, 1991, pp. 100 -- 109. [Online]. Available:
  \url{https://doi.org/10.1117/12.49324}
\BIBentrySTDinterwordspacing

\bibitem{Tendero12}
\BIBentryALTinterwordspacing
Y.~Tendero and J.~Gilles, ``Admire: a locally adaptive single-image,
  non-uniformity correction and denoising algorithm: application to uncooled ir
  camera,'' \emph{Infrared Technology and Applications}, vol. 8353, pp.
  580--595, 5 2012. [Online]. Available:
  \url{https://doi.org/10.1117/12.912966}
\BIBentrySTDinterwordspacing

\bibitem{Cao2014}
\BIBentryALTinterwordspacing
Y.~Cao and C.-L. Tisse, ``Single-image-based solution for optics
  temperature-dependent nonuniformity correction in an uncooled long-wave
  infrared camera,'' \emph{Optics Letters}, vol.~39, no.~3, p. 646, Jan. 2014.
  [Online]. Available: \url{https://doi.org/10.1364/ol.39.000646}
\BIBentrySTDinterwordspacing

\bibitem{Zhao13}
J.~Zhao, Q.~Zhou, Y.~Chen, T.~Liu, H.~Feng, Z.~Xu, and Q.~Li, ``Single image
  stripe nonuniformity correction with gradient-constrained optimization model
  for infrared focal plane arrays,'' \emph{Optics Communications}, vol. 296,
  pp. 47--52, 2013.

\bibitem{Jian2018}
X.~Jian, C.~Lv, and R.~Wang, ``Nonuniformity correction of single infrared
  images based on deep filter neural network,'' \emph{Symmetry}, 2018.

\bibitem{He2018}
Z.~He, Y.~Y. Cao, Y.~Dong, J.~Yang, and C.-L. Tisse, ``Single-image-based
  nonuniformity correction of uncooled long-wave infrared detectors: a
  deep-learning approach,'' \emph{Applied Optics}, vol.~57, p. D155, 2018.

\bibitem{ChangDeepLearning2019}
Y.~Chang, L.~Yan, L.~Liu, H.~Fang, and S.~Zhong, ``Infrared aerothermal
  nonuniform correction via deep multiscale residual network,'' \emph{IEEE
  Geoscience and Remote Sensing Letters}, vol.~16, pp. 1120--1124, 2019,
  https://owuchangyuo.github.io/files/DMRN.rar<br/>Code.

\bibitem{Saragadam2021}
\BIBentryALTinterwordspacing
V.~Saragadam, A.~Dave, A.~Veeraraghavan, and R.~G. Baraniuk, ``Thermal image
  processing via physics-inspired deep networks,'' in \emph{IEEE Conference on
  Computer Vision and Pattern Recognition (CVPR)}, 2021. [Online]. Available:
  \url{https://github.com/vishwa91/DeepIR}
\BIBentrySTDinterwordspacing

\bibitem{Oz2022}
N.~Oz, N.~Sochen, D.~Mendelovich, and I.~Klapp, ``Improving temperature
  estimation in low-cost infrared cameras using deep neural networks,''
  \emph{Arxiv}, 2022.

\bibitem{Harris99}
J.~Harris and Y.-M. Chiang, ``Nonuniformity correction of infrared image
  sequences using the constant-statistics constraint,'' \emph{IEEE Transactions
  on Image Processing}, vol.~8, no.~8, pp. 1148--1151, 1999.

\bibitem{Hardie00}
\BIBentryALTinterwordspacing
R.~C. Hardie, M.~M. Hayat, E.~Armstrong, and B.~Yasuda, ``Scene-based
  nonuniformity correction with video sequences and registration,'' \emph{Appl.
  Opt.}, vol.~39, no.~8, pp. 1241--1250, Mar 2000. [Online]. Available:
  \url{https://opg.optica.org/ao/abstract.cfm?URI=ao-39-8-1241}
\BIBentrySTDinterwordspacing

\bibitem{Vera05}
E.~Vera and S.~Torres, ``Fast adaptive nonuniformity correction for infrared
  focal-plane array detectors,'' \emph{EURASIP Journal on Applied Signal
  Processing}, 2005.

\bibitem{Averbuch2007}
A.~Averbuch, G.~Liron, and B.~Z. Bobrovsky, ``Scene based non-uniformity
  correction in thermal images using kalman filter,'' \emph{Image and Vision
  Computing}, vol.~25, pp. 833--851, 6 2007.

\bibitem{Zuo11}
\BIBentryALTinterwordspacing
C.~Zuo, Q.~Chen, G.~Gu, and X.~Sui, ``Scene-based nonuniformity correction
  algorithm based on interframe registration,'' \emph{J. Opt. Soc. Am. A},
  vol.~28, no.~6, pp. 1164--1176, Jun 2011. [Online]. Available:
  \url{https://opg.optica.org/josaa/abstract.cfm?URI=josaa-28-6-1164}
\BIBentrySTDinterwordspacing

\bibitem{Papini2018}
S.~Papini, P.~Yafin, I.~Klapp, and N.~Sochen, ``Joint estimation of unknown
  radiometric data, gain, and offset from thermal images,'' \emph{Applied
  Optics}, vol.~57, p. 10390, 2018.

\bibitem{uncooled_thermal_imaging}
P.~W. Kruse, \emph{Uncooled thermal imaging arrays, systems, and
  applications}.\hskip 1em plus 0.5em minus 0.4em\relax SPIE, 2001.

\bibitem{ir_1f_gaussian}
\BIBentryALTinterwordspacing
R.~F. Voss, ``Linearity of $\frac{1}{f}$ noise mechanisms,'' \emph{Phys. Rev.
  Lett.}, vol.~40, pp. 913--916, Apr 1978. [Online]. Available:
  \url{https://link.aps.org/doi/10.1103/PhysRevLett.40.913}
\BIBentrySTDinterwordspacing

\bibitem{computationalGeometry}
M.~d. Berg, O.~Cheong, M.~v. Kreveld, and M.~Overmars, \emph{Computational
  Geometry: Algorithms and Applications}, 3rd~ed.\hskip 1em plus 0.5em minus
  0.4em\relax Santa Clara, CA, USA: Springer-Verlag TELOS, 2008.

\bibitem{HartleyRichard2004MVGi}
R.~Hartley and A.~Zisserman, \emph{\BIBforeignlanguage{eng}{Multiple View
  Geometry in Computer Vision}}.\hskip 1em plus 0.5em minus 0.4em\relax
  Cambridge: Cambridge University Press, 2004.

\bibitem{multi_sr_classic}
S.~Farsiu, M.~Robinson, M.~Elad, and P.~Milanfar, ``Fast and robust multiframe
  super resolution,'' \emph{IEEE Transactions on Image Processing}, vol.~13,
  no.~10, pp. 1327--1344, 2004.

\bibitem{multi_sr_classic2}
S.~Kim and W.-Y. Su, ``Recursive high-resolution reconstruction of blurred
  multiframe images,'' in \emph{[Proceedings] ICASSP 91: 1991 International
  Conference on Acoustics, Speech, and Signal Processing}, 1991, pp. 2977--2980
  vol.4.

\bibitem{multi_denoise_classic}
M.~Zhang and B.~K. Gunturk, ``Multiresolution bilateral filtering for image
  denoising,'' \emph{IEEE Transactions on Image Processing}, vol.~17, no.~12,
  pp. 2324--2333, 2008.

\bibitem{multi_deblur_classic}
O.~Whyte, J.~Sivic, A.~Zisserman, and J.~Ponce, ``Non-uniform deblurring for
  shaken images,'' \emph{International Journal of Computer Vision}, vol.~98,
  pp. 168--186, 6 2012.

\bibitem{Detone16}
\BIBentryALTinterwordspacing
D.~DeTone, T.~Malisiewicz, and A.~Rabinovich, ``Deep image homography
  estimation,'' 2016. [Online]. Available:
  \url{https://arxiv.org/abs/1606.03798}
\BIBentrySTDinterwordspacing

\bibitem{multi_denoise_dl}
C.~Godard, K.~Matzen, and M.~Uyttendaele, ``Deep burst denoising,'' in
  \emph{15th European Conference of Computer Vision ECCV}, 2018.

\bibitem{multi_sr_dl1}
\BIBentryALTinterwordspacing
G.~Bhat, M.~Danelljan, L.~V. Gool, and R.~Timofte, ``Deep burst
  super-resolution,'' in \emph{IEEE/CVF Conference on Computer Vision and
  Pattern Recognition (CVPR)}.\hskip 1em plus 0.5em minus 0.4em\relax Los
  Alamitos, CA, USA: IEEE Computer Society, jun 2021, pp. 9205--9214. [Online].
  Available:
  \url{https://doi.ieeecomputersociety.org/10.1109/CVPR46437.2021.00909}
\BIBentrySTDinterwordspacing

\bibitem{multi_sr_dl2}
\BIBentryALTinterwordspacing
B.~Wronski, I.~Garcia-Dorado, M.~Ernst, D.~Kelly, M.~Krainin, C.-K. Liang,
  M.~Levoy, and P.~Milanfar, ``Handheld multi-frame super-resolution,''
  \emph{ACM Trans. Graph.}, vol.~38, no.~4, jul 2019. [Online]. Available:
  \url{https://doi.org/10.1145/3306346.3323024}
\BIBentrySTDinterwordspacing

\bibitem{multi_sr_dl3}
M.~Deudon, A.~Kalaitzis, I.~Goytom, M.~R. Arefin, Z.~Lin, K.~Sankaran,
  V.~Michalski, S.~E. Kahou, J.~Cornebise, and Y.~Bengio, ``Highres-net:
  Recursive fusion for multi-frame super-resolution of satellite imagery,''
  \emph{ArXiv}, vol. abs/2002.06460, 2020.

\bibitem{DeBrabandere16}
B.~De~Brabandere, X.~Jia, T.~Tuytelaars, and L.~Van~Gool, ``Dynamic filter
  networks,'' in \emph{30th International Conference on Neural Information
  Processing Systems (NIPS)}, 2016.

\bibitem{unet}
O.~Ronneberger, P.~Fischer, and T.~Brox, ``U-net: Convolutional networks for
  biomedical image segmentation,'' in \emph{Medical Image Computing and
  Computer-Assisted Intervention -- MICCAI 2015}, N.~Navab, J.~Hornegger, W.~M.
  Wells, and A.~F. Frangi, Eds.\hskip 1em plus 0.5em minus 0.4em\relax Cham:
  Springer International Publishing, 2015, pp. 234--241.

\bibitem{ssimLoss2017}
H.~Zhao, O.~Gallo, I.~Frosio, and J.~Kautz, ``Loss functions for image
  restoration with neural networks,'' \emph{IEEE Transactions on Computational
  Imaging}, vol.~3, no.~1, pp. 47--57, 2017.

\bibitem{mildenhall2018kpn}
B.~Mildenhall, J.~T. Barron, J.~Chen, D.~Sharlet, R.~Ng, and R.~Carroll,
  ``Burst denoising with kernel prediction networks,'' in \emph{IEEE Conference
  on Computer Vision and Pattern Recognition (CVPR)}, 2018.

\bibitem{anwar2020}
\BIBentryALTinterwordspacing
S.~Anwar, S.~Khan, and N.~Barnes, ``A deep journey into super-resolution: A
  survey,'' \emph{ACM Comput. Surv.}, vol.~53, no.~3, may 2020. [Online].
  Available: \url{https://doi.org/10.1145/3390462}
\BIBentrySTDinterwordspacing

\bibitem{python}
G.~Van~Rossum and F.~L. Drake, \emph{Python 3 Reference Manual}.\hskip 1em plus
  0.5em minus 0.4em\relax Scotts Valley, CA: CreateSpace, 2009.

\bibitem{pytorch}
A.~Paszke, S.~Gross, S.~Chintala, G.~Chanan, E.~Yang, Z.~DeVito, Z.~Lin,
  A.~Desmaison, L.~Antiga, and A.~Lerer, ``Automatic differentiation in
  pytorch,'' \emph{NIPS}, 2017.

\bibitem{adamOpt2015}
\BIBentryALTinterwordspacing
D.~P. Kingma and J.~Ba, ``Adam: {A} method for stochastic optimization,'' in
  \emph{3rd International Conference on Learning Representations, {ICLR} 2015,
  San Diego, CA, USA, May 7-9, 2015, Conference Track Proceedings}, Y.~Bengio
  and Y.~LeCun, Eds., 2015. [Online]. Available:
  \url{http://arxiv.org/abs/1412.6980}
\BIBentrySTDinterwordspacing

\bibitem{orthoInit2014}
\BIBentryALTinterwordspacing
A.~M. Saxe, J.~L. McClelland, and S.~Ganguli, ``Exact solutions to the
  nonlinear dynamics of learning in deep linear neural networks,'' in \emph{2nd
  International Conference on Learning Representations, {ICLR} 2014, Banff, AB,
  Canada, April 14-16, 2014, Conference Track Proceedings}, Y.~Bengio and
  Y.~LeCun, Eds., 2014. [Online]. Available:
  \url{http://arxiv.org/abs/1312.6120}
\BIBentrySTDinterwordspacing

\bibitem{gelu}
D.~Hendrycks and K.~Gimpel, ``Gaussian error linear units (gelus),''
  \emph{arXiv preprint arXiv:1606.08415}, 2016.

\bibitem{batchNorm}
\BIBentryALTinterwordspacing
S.~Ioffe and C.~Szegedy, ``Batch normalization: Accelerating deep network
  training by reducing internal covariate shift,'' in \emph{Proceedings of the
  32nd International Conference on Machine Learning}, ser. Proceedings of
  Machine Learning Research, F.~Bach and D.~Blei, Eds., vol.~37.\hskip 1em plus
  0.5em minus 0.4em\relax Lille, France: PMLR, 07--09 Jul 2015, pp. 448--456.
  [Online]. Available: \url{https://proceedings.mlr.press/v37/ioffe15.html}
\BIBentrySTDinterwordspacing

\bibitem{shuffle_block}
W.~Shi, J.~Caballero, F.~Husz{\'a}r, J.~Totz, A.~P. Aitken, R.~Bishop,
  D.~Rueckert, and Z.~Wang, ``Real-time single image and video super-resolution
  using an efficient sub-pixel convolutional neural network,'' \emph{2016 IEEE
  Conference on Computer Vision and Pattern Recognition (CVPR)}, pp.
  1874--1883, 2016.

\end{thebibliography}

\clearpage
\appendix

\newcommand{\heightDiffResSupp}{13ex}
\newcommand{\makeRowResultsSupp}[1]{\subfloat{\includegraphics[height=\heightDiffResSupp]{res_diff/#1/rect.pdf}}
                                \hfill
                                \subfloat{\includegraphics[height=\heightDiffResSupp]{res_diff/#1/diff/7.pdf}}
                                \hfill
                                \subfloat{\includegraphics[height=\heightDiffResSupp]{res_diff/#1/diff/9.pdf}}
                                \hfill
                                \subfloat{\includegraphics[height=\heightDiffResSupp]{res_diff/#1/diff/11.pdf}}
                                \hfill
                                \subfloat{\includegraphics[height=\heightDiffResSupp]{res_diff/#1/cbar.pdf}}}
\newcommand{\makeResultsFigSupp}[7]{\begin{figure*}
                                    \centering
                                    \begin{minipage}{0.49\linewidth}
                                        \makeRowResultsSupp{#3}
                                    \end{minipage}\hfill
                                    \begin{minipage}{0.49\linewidth}
                                        \makeRowResultsSupp{#4}
                                    \end{minipage}
                                    \\
                                    \begin{minipage}{0.49\linewidth}
                                        \makeRowResultsSupp{#5}
                                    \end{minipage}\hfill
                                    \begin{minipage}{0.49\linewidth}
                                        \makeRowResultsSupp{#6}
                                    \end{minipage}
                                    \\
                                    \begin{minipage}{0.49\linewidth}
                                        \setcounter{subfigure}{0}
                                        \subfloat[GT]{\includegraphics[height=\heightDiffResSupp]{res_diff/#1/rect.pdf}}
                                        \hfill
                                        \subfloat[7]{\includegraphics[height=\heightDiffResSupp]{res_diff/#1/diff/7.pdf}}
                                        \hfill
                                        \subfloat[9]{\includegraphics[height=\heightDiffResSupp]{res_diff/#1/diff/9.pdf}}
                                        \hfill
                                        \subfloat[11]{\includegraphics[height=\heightDiffResSupp]{res_diff/#1/diff/11.pdf}}
                                        \hfill
                                        \subfloat{\includegraphics[height=\heightDiffResSupp]{res_diff/#1/cbar.pdf}}
                                    \end{minipage}\hfill
                                    \begin{minipage}{0.49\linewidth}
                                        \setcounter{subfigure}{0}
                                        \subfloat[GT]{\includegraphics[height=\heightDiffResSupp]{res_diff/#2/rect.pdf}}
                                        \hfill
                                        \subfloat[7]{\includegraphics[height=\heightDiffResSupp]{res_diff/#2/diff/7.pdf}}
                                        \hfill
                                        \subfloat[9]{\includegraphics[height=\heightDiffResSupp]{res_diff/#2/diff/9.pdf}}
                                        \hfill
                                        \subfloat[11]{\includegraphics[height=\heightDiffResSupp]{res_diff/#2/diff/11.pdf}}
                                        \hfill
                                        \subfloat{\includegraphics[height=\heightDiffResSupp]{res_diff/#2/cbar.pdf}}
                                    \end{minipage}
                                    \caption{Difference between the temperature estimation with our method and the ground truth. The left-most figure is the ground truth. The following figures are the zoom-in of the area inside the red rectangle. The number bellow the difference map is the number of frames used for the temperature estimation, from left to right 7, 9 and 11 frames.}
                                    \label{supp_diff_#7}
                                \end{figure*}
}
\newcommand{\heightPatchResSupp}{0.16\linewidth}%
\newcommand{\makePatchRowSupp}[1]{
    \subfloat{\includegraphics[height=\heightPatchResSupp]{res_patch/#1/sample.pdf}}
    \hfill
    \subfloat{\includegraphics[height=\heightPatchResSupp]{res_patch/#1/label.pdf}}
    \hfill
    \subfloat{\includegraphics[height=\heightPatchResSupp]{res_patch/#1/Ours.pdf}}
    \hfill
    \subfloat{\includegraphics[height=\heightPatchResSupp]{res_patch/#1/ADMIRE.pdf}}
    \hfill
    \subfloat{\includegraphics[height=\heightPatchResSupp]{res_patch/#1/DeepIR.pdf}}
    \hfill
    \subfloat{\includegraphics[height=\heightPatchResSupp]{res_patch/#1/He_et_al.pdf}}
}%
\newcommand{\makePatchesFigSupp}[7]{\begin{figure*}
        \centering
        \makePatchRowSupp{#1}\\
        \makePatchRowSupp{#2}\\
        \makePatchRowSupp{#3}\\
        \makePatchRowSupp{#4}\\
        \makePatchRowSupp{#5}\\
        \setcounter{subfigure}{0}
        \subfloat[Sample]{\includegraphics[height=\heightPatchResSupp]{res_patch/#6/sample.pdf}}
        \hfill
        \subfloat[GT]{\includegraphics[height=\heightPatchResSupp]{res_patch/#6/label.pdf}}
        \hfill
        \subfloat[Ours]{\includegraphics[height=\heightPatchResSupp]{res_patch/#6/Ours.pdf}}
        \hfill
        \subfloat[ADMIRE~\cite{Tendero12}]{\includegraphics[height=\heightPatchResSupp]{res_patch/#6/ADMIRE.pdf}}
        \hfill
        \subfloat[DeepIR~\cite{Saragadam2021}]{\includegraphics[height=\heightPatchResSupp]{res_patch/#6/DeepIR.pdf}}
        \hfill
        \subfloat[He~\cite{He2018}]{\includegraphics[height=\heightPatchResSupp]{res_patch/#6/He_et_al.pdf}}
        \caption{Zoomed-in results of different methods.
        The left-most figure is the reference frame with a red rectangle. 
        The following figures are the results of the area inside the red rectangle.
        $\nFrames=11$ for all results.}
        \label{supp_patch_#7}
    \end{figure*}}%
\newcommand{\makePatchesFigHallucinationSupp}[7]{\begin{figure*}
        \centering
        \makePatchRowSupp{#1}\\
        \makePatchRowSupp{#2}\\
        \makePatchRowSupp{#3}\\
        \makePatchRowSupp{#4}\\
        \makePatchRowSupp{#5}\\
        \setcounter{subfigure}{0}
        \subfloat[Sample]{\includegraphics[height=\heightPatchResSupp]{res_patch/#6/sample.pdf}}
        \hfill
        \subfloat[GT]{\includegraphics[height=\heightPatchResSupp]{res_patch/#6/label.pdf}}
        \hfill
        \subfloat[Ours]{\includegraphics[height=\heightPatchResSupp]{res_patch/#6/Ours.pdf}}
        \hfill
        \subfloat[ADMIRE~\cite{Tendero12}]{\includegraphics[height=\heightPatchResSupp]{res_patch/#6/ADMIRE.pdf}}
        \hfill
        \subfloat[DeepIR~\cite{Saragadam2021}]{\includegraphics[height=\heightPatchResSupp]{res_patch/#6/DeepIR.pdf}}
        \hfill
        \subfloat[He~\cite{He2018}]{\includegraphics[height=\heightPatchResSupp]{res_patch/#6/He_et_al.pdf}}
        \caption{Zoomed-in results of different methods.
        The left-most figure is the reference frame with a red rectangle. 
        The following figures are the results of the area inside the red rectangle.
        $\nFrames=11$ for all results.
        These results displays the hallucination effect of DeepIR~\cite{Saragadam2021} method.}
        \label{supp_patch_#7}
    \end{figure*}}%

\section*{Network Architecture}
\begin{figure*}
    \centering
    \includegraphics[width=\linewidth]{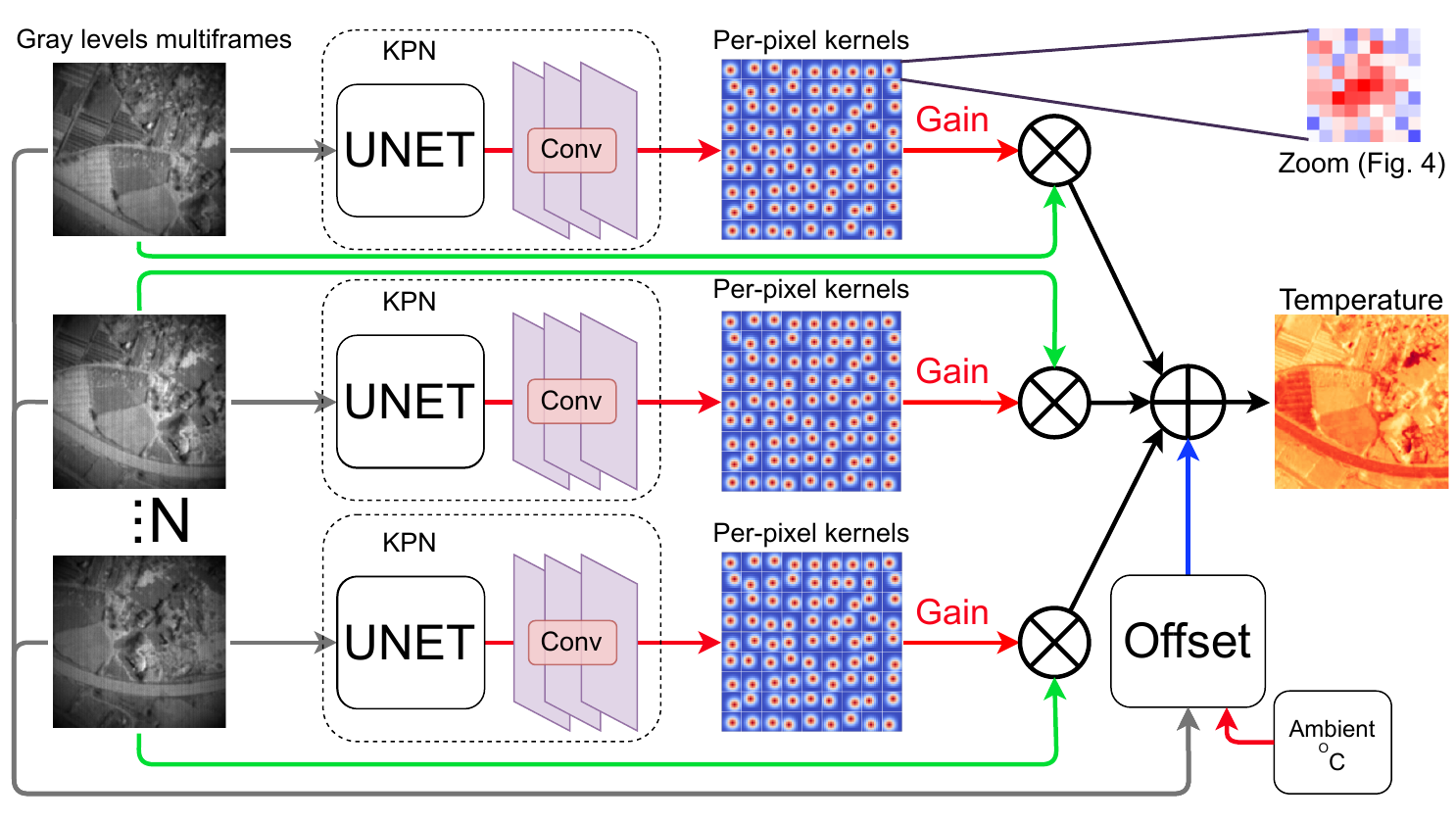}
    \caption{Schematics of the model.
        The gray-level multiframes are fed into the kernel prediction network (KPN), and the KPN outputs the per-pixel kernels for each frame.
        Each frame is divided to overlapping patches with the same support as the kernels. 
        The patches and the kernels are multiplied element-wise and each product is summed, resulting in a 2D gain map for each frame.
        All the 2D gain maps are summed depth-wise, resulting in a single 2D map.
        The offset, a single scalar value, is added to the single 2D map to get the estimated temperature map.}
    \label{supp_network}
\end{figure*}

A schematic diagram of the whole network is given in \cref{fig:methods:network} in the main text and enlarged in \cref{supp_network}. Below is a detailed description of the network architecture, from the UNET encoder-decoder to the kernel estimation block and the offset block.

First, we describe the UNET encoder-decoder.
We use a tensor with $\nFrames$ channels of gray-level frames as the input to the network. The input tensor undergoes a $3\times 3$ convolution (conv) layer that encodes it from $\nFrames$ channels to $\nChannels$ channels without any activation function. The encoded features then pass through the encoder and decoder parts of the network, where the number of channels is multiplied by a factor of $\scaleFactor$ at each level. The encoder and decoder blocks consist of three $3\times 3$ conv-layers each. The first two layers in each block have GeLU~\cite{gelu} and batch normalization (norm)~\cite{batchNorm}, while the last layer has neither activation nor norm. The last layer in each block produces $\nChannels\times\scaleFactor^i$ channels, where $i$ is the level index. The encoder block also applies an average pooling layer with a $\scaleFactor\times\scaleFactor$ window and stride $\scaleFactor$ to reduce the spatial resolution, while the decoder block uses a pixel shuffle layer~\cite{shuffle_block} with an upsample factor of $\scaleFactor$ to increase it. We concatenate the encoder block output before pooling with the pixel shuffle output at each level to feed it into the decoder block. The encoder and decoder block structures are shown in \cref{table:nn_architecture} in the main text.

After the last decoder block in the UNET, we add a kernel-estimation block composed of three $1\times1$ conv layers. The first two layers have GeLU, and the last layer has no activation. The kernel estimation block structure is shown in \cref{table:kpnBlock} in the main text. 
The block outputs $\nFrames\times K\times K$ channels. We reshape the output of this block as kernels of size $K\times K$ for each frame. We then sample a patch of size $K\times K$ around each pixel in each frame and compute the inner product of the corresponding kernel and patch, as in the first term of \cref{eq:methods:KPN} in the main text. We sum the inner products from all frames for each pixel.

To map the temperature estimation to the camera range, we use an offset block that takes the means of all input gray-level frames as input and outputs a single scalar. 
The offset block is a fully connected layer that acts as a polynomial function of the input. The offset block is explained in more detail in \cref{sec:methods:net} in the main text.

The final temperature estimation is obtained by adding the offset scalar to the pixel-wise summation of the gain from the kernel estimation block.

The scale factor for decoder and encoder blocks is $\scaleFactor\equiv2$ and the number of channels is $\nChannels\equiv64$ throughout the work. 
The number of levels was empirically set to $4$.

\begin{table*}[h]
    \caption{Architecture of the encoder and decoder blocks.}
    \centering
    \begin{tabular}{c c c c c}
        \hline
        \textbf{Type} & \textbf{BN}~\cite{batchNorm} & \textbf{GeLU}~\cite{gelu} & \textbf{Kernel} & \textbf{Output} \\ [0.5ex]
        \hline
        Conv2D & $\surd$ & $\surd$ & 3 & $\nChannels \times h \times w$ \\
        Conv2D & $\surd$ & $\surd$ & 3 & $\nChannels \times h \times w$ \\
        Conv2D & $\times$ & $\times$ & 3 & $(\scaleFactor \times \nChannels) \times h \times w$ \\
        \hline
    \end{tabular}
    \label{table:nn_architecture}
\end{table*}
\begin{table*}[h]
    \caption{Architecture of the kernel predictor block.}
    \centering
        \begin{tabular}{c c c c}
            \hline
            \textbf{Type} &  \textbf{GeLU}~\cite{gelu} & \textbf{Kernel} & \textbf{Output} \\ [0.5ex]
            \hline
            Conv2D & $\surd$ & 1 & $\nChannels \times h \times w$ \\
            Conv2D & $\surd$ & 1 & $\nChannels \times h \times w$ \\
            Conv2D & $\times$ & 1 & $ (\nFrames\times K^2) \times h \times w$ \\
            \hline
        \end{tabular}
    \label{table:kpnBlock}
\end{table*}

\section*{Figure list}
\begin{figure*}
    \centering
    \includegraphics[width=0.9\linewidth]{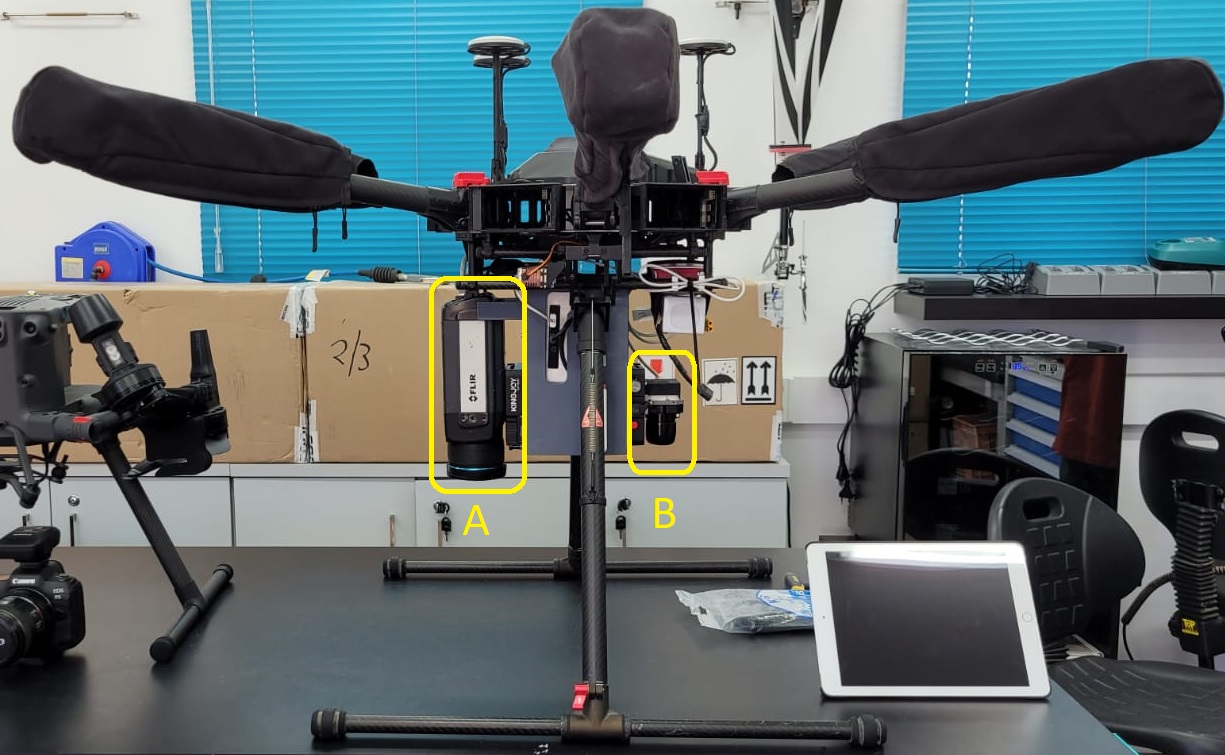}
    \caption{UAV used for the collection of real data. \scientificCamera\ is mounted on the left (marked A) and \taucamera\ is mounted on the right (marked B).}
    \label{supp_uav}
\end{figure*}

\begin{enumerate}
    \item \cref{supp_realdata} shows more results of the proposed method on real data.
    \item \cref{supp_diff_7,supp_diff_8,supp_diff_9,supp_diff_10,supp_diff_11} display the mean absolute error per pixel as a function of number of frames, both quantitatively and qualitatively.
    \item \cref{supp_patch_1,supp_patch_2,supp_patch_3,supp_patch_4,supp_patch_5,supp_patch_6} compare the results of the proposed method to ADMIRE~\cite{Tendero12}, DeepIR~\cite{Saragadam2021} and He~\cite{He2018} et al.'s methods. \cref{supp_patch_5,supp_patch_6} specifically show the hallucination effect of the DeepIR~\cite{Saragadam2021} method.
    \item \cref{supp_H,supp_I,supp_A,supp_O,supp_B,supp_M} are the original images used for the real data results in \cref{fig:results:realdata} in the main text. On the left of each figure is the ground truth (GT) temperature map acquired by the \scientificCamera, and on the right is the temperature map estimated by the proposed method. The raw data cannot be displayed because they consist of 7 frames. Both the GT and the estimated temperature map undergoes an histogram equalization to improve the visualization.
    \item \cref{supp_uav} shows the unmanned aerial vehicle (UAV) used for the real data experiments.
    \item \cref{tab:cameraParams} specifies the \taucamera\ parameters used throughout all of the experiments.
\end{enumerate}

\newcommand{\sizeRealDataSupp}{0.48}%
\begin{figure*}
    \centering
    \subfloat[]{\includegraphics[width=\sizeRealDataSupp\linewidth]{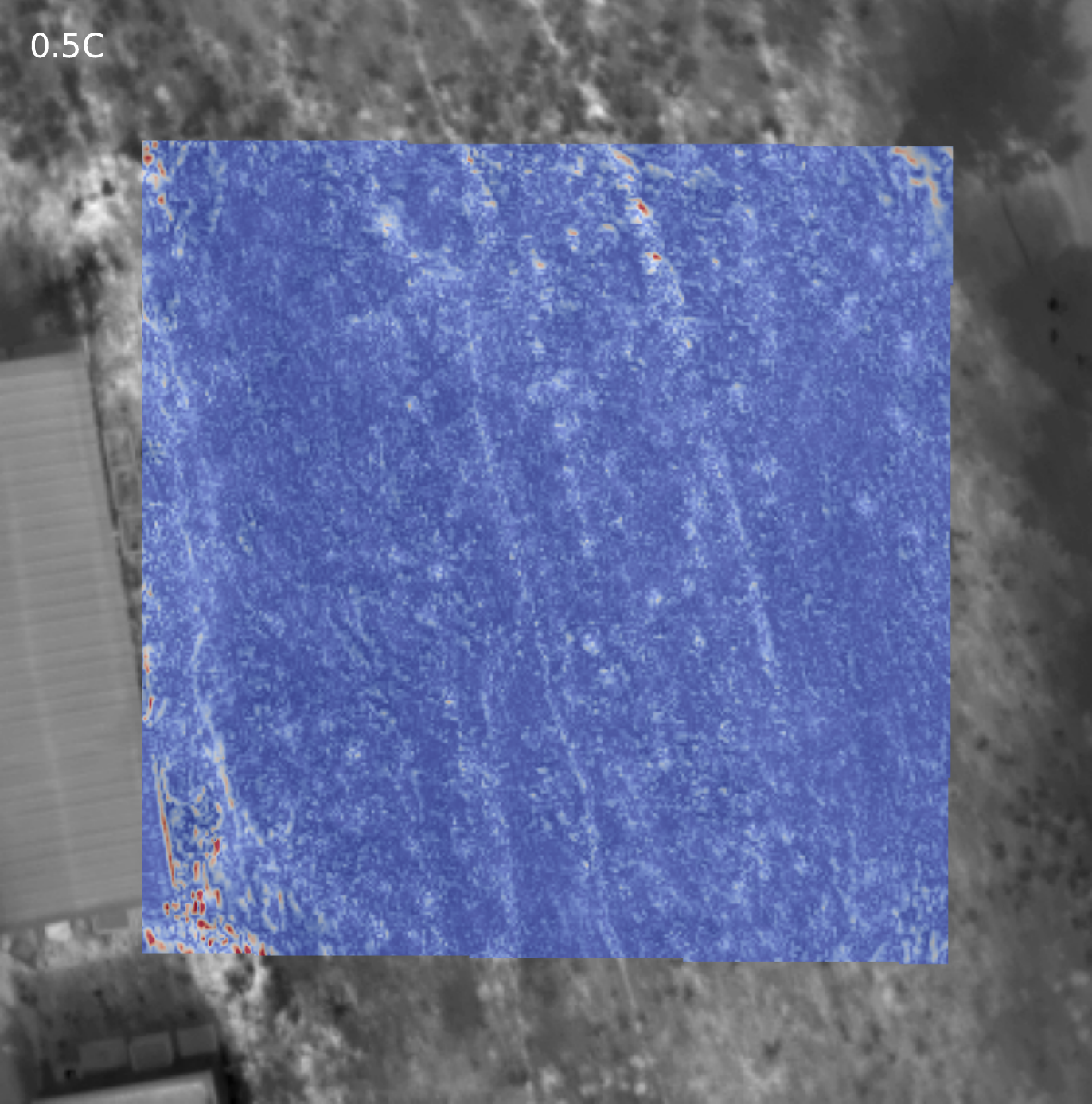}}
    \hfill
    \subfloat[]{\includegraphics[width=\sizeRealDataSupp\linewidth]{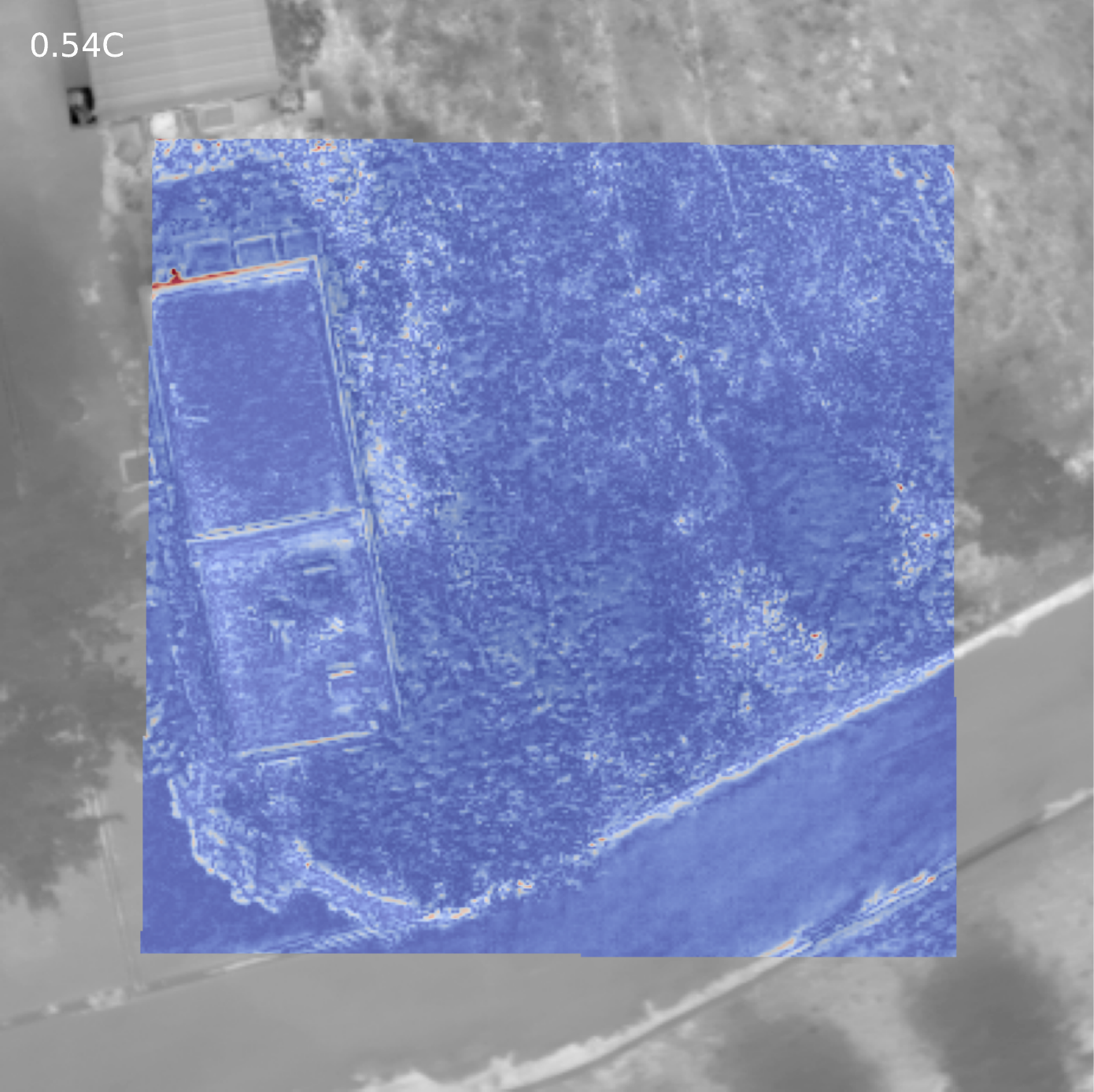}}
    \\
    \subfloat[]{\includegraphics[width=\sizeRealDataSupp\linewidth]{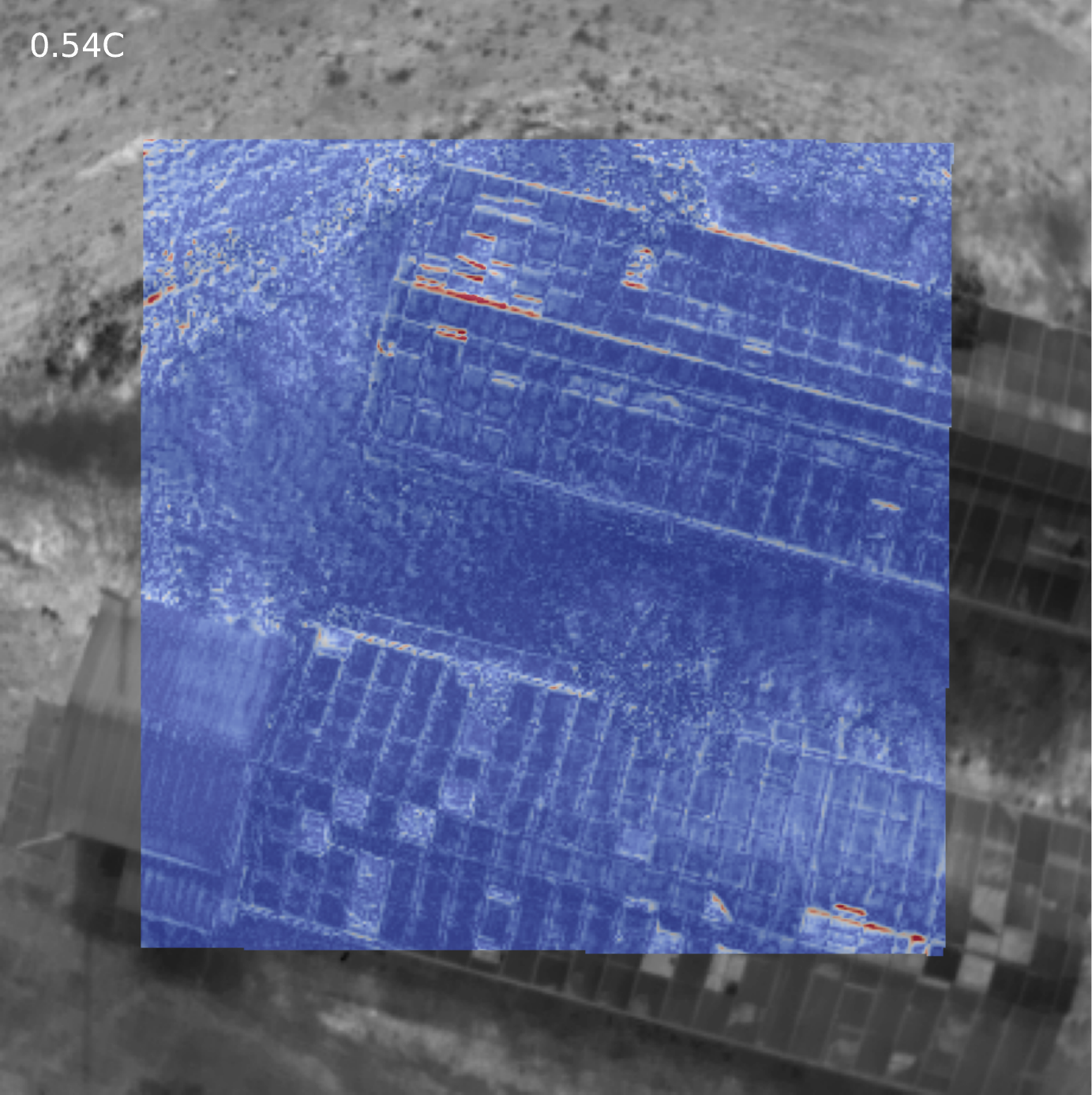}}
    \hfill
    \subfloat[]{\includegraphics[width=\sizeRealDataSupp\linewidth]{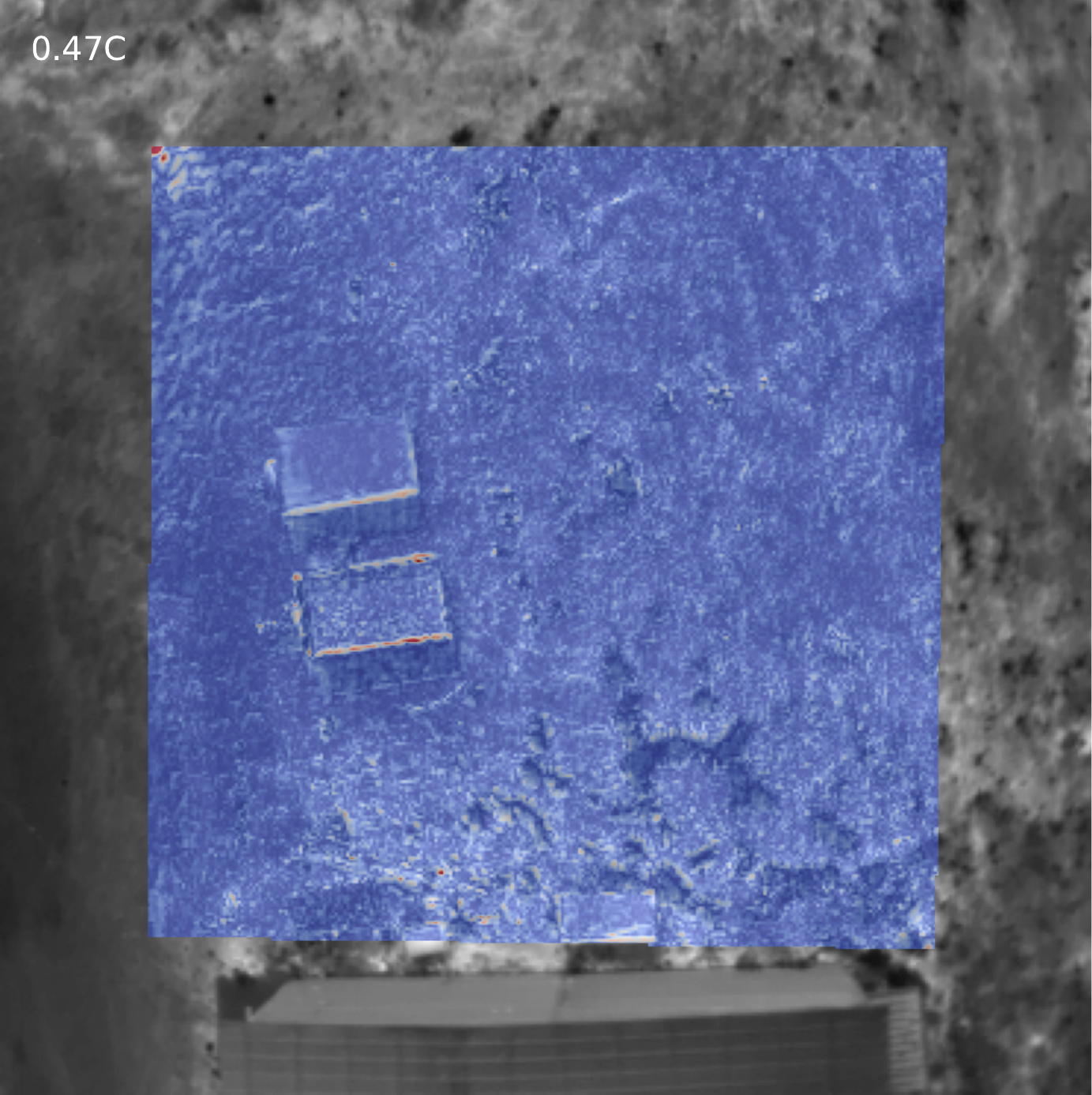}}
    \caption{Various results on real data. The gray background is the GT temperature map and the colored map is the difference between the GT and estimated temperature maps. The number on the top-left corner is the MAE between the estimated and GT temperature maps.}
    \label{supp_realdata}
\end{figure*}%

\makeResultsFigSupp{180805_Peach_18}{Gilat_210809_0}{Gilat_210809_3}{Gilat_210809_4}{Gilat_210809_11}{Gilat_210809_18}{7}
\makeResultsFigSupp{Gilat_210809_19}{MevoBytar_210818_0}{MevoBytar_210818_3}{MevoBytar_210818_7}{MevoBytar_210818_8}{MevoBytar_210818_9}{8}
\makeResultsFigSupp{MevoBytar_210818_11}{MevoBytar_210818_13}{NeveYaar_210520_0}{NeveYaar_210520_5}{NeveYaar_210520_6}{NeveYaar_210520_9}{9}
\makeResultsFigSupp{NeveYaar_210520_14}{NirEliyho_211005_5}{NirEliyho_211005_9}{NirEliyho_211005_14}{NirEliyho_211005_16}{Tzora_210523_3}{10}
\makeResultsFigSupp{Tzora_210523_8}{Tzora_210523_14}{Tzora_210523_16}{YanivReshef_190816_7}{YanivReshef_190816_8}{YanivReshef_190816_11}{11}

\makePatchesFigSupp{180725_Ramon_2}{180725_Ramon_4}{180725_Ramon_5}{180725_Ramon_7}{180725_Ramon_8}{180725_Ramon_10}{1}%
\makePatchesFigSupp{180725_Ramon_11}{180805_Peach_14}{180805_Peach_16}{180805_Peach_34}{180805_Peach_35}{180805_Peach_40}{2}%
\makePatchesFigSupp{180805_Peach_57}{180805_Peach_60}{180805_Peach_87}{180805_Peach_90}{180805_Peach_91}{Gilat_210809_37}{3}%
\makePatchesFigSupp{Gilat_210809_44}{Gilat_210809_260}{MevoBytar_210818_42}{MevoBytar_210818_82}{MevoBytar_210818_96}{NirEliyho_211005_560}{4}%
\makePatchesFigHallucinationSupp{NeveYaar_210520_14}{NeveYaar_210520_55}{NeveYaar_210520_148}{NeveYaar_210520_394}{NeveYaar_210520_421}{NeveYaar_210520_424}{5}%
\makePatchesFigHallucinationSupp{NeveYaar_210520_867}{NeveYaar_210520_966}{NeveYaar_210520_1033}{NeveYaar_210520_867}{NirEliyho_211005_169}{Tzora_210523_35}{6}%

\newcommand{\sizeGTSupp}{0.48}%
\newcommand{\sizeEstSupp}{0.36}%
\newcommand{\makeGTSupp}[3]{
    \begin{figure*}
        \centering
        \subfloat[]{\includegraphics[width=\sizeGTSupp\linewidth]{real_data/main/#1/gt.pdf}}
        \hfill
        \subfloat[]{\includegraphics[width=\sizeEstSupp\linewidth]{real_data/main/#1/est.pdf}}
        \caption{Ground truth (left) and estimated (right) temperature maps for the result in #2.}
        \label{supp_#1}
    \end{figure*}
}
\makeGTSupp{H}{\cref{fig:results:realdata} (a)}\\
\makeGTSupp{I}{\cref{fig:results:realdata} (b)}\\
\makeGTSupp{A}{\cref{fig:results:realdata} (c)}\\
\makeGTSupp{O}{\cref{fig:results:realdata} (d)}\\
\makeGTSupp{B}{\cref{fig:results:realdata} (e)}\\
\makeGTSupp{M}{\cref{fig:results:realdata} (f)}

\begin{table*}[h]
    \centering
    \caption{The FLIR \taucamera\ settings as described in Tau2 Quark Software IDD}
    \begin{tabular}{|c|c||c|c|}
        \hline
        Function   & State  & Function        & State                  \\
        \hline\hline
        FFC Mode   & Auto   & FPS             & 4 ($60_{Hz}$)          \\
        \hline
        FFC Period & 0      & CMOS Depth      & 0 ($14_{bit}$ w/o AGC) \\
        \hline
        Isotherm   & 0      & LVDS            & 0                      \\
        \hline
        DDE        & 0      & LVDS Depth      & 0 ($14_{bit}$)         \\
        \hline
        T-Linear   & 0      & XP              & 2 ($14_{bit}$)         \\
        \hline
        AGC        & Manual & Brightness Bias & 0                      \\
        \hline
        Contrast   & 0      & Brightness      & 0                      \\
        \hline
        ACE        & 0      & SSO             & 0                      \\
        \hline
        Gain       & High   &                 &                        \\
        \hline
    \end{tabular}
    \label{tab:cameraParams}
\end{table*}

\end{document}